\documentclass{article}
\pdfoutput=1

\usepackage{adjustbox}

\usepackage{textcomp}

\usepackage[hidelinks]{hyperref}

\usepackage[absolute]{textpos}
\usepackage{graphicx}

\usepackage[normalem]{ulem}

\usepackage[utf8]{inputenc}
\usepackage[T1]{fontenc}
\usepackage{lmodern}
\usepackage{caption}

\def\independenT#1#2{\mathrel{\rlap{$#1#2$}\mkern2mu{#1#2}}}
\newcommand\indep{\protect\mathpalette{\protect\independenT}{\perp}}

\makeatletter
\newcommand{\xdashrightarrow}[2][]{\ext@arrow 0359\rightarrowfill@@{#1}{#2}}
\def\rightarrowfill@@{\arrowfill@@\relax\relbar\rightarrow}
\makeatother
\makeatletter
\newcommand{\xdashleftarrow}[2][]{\ext@arrow 0359\leftarrowfill@@{#1}{#2}}
\def\leftarrowfill@@{\arrowfill@@\relax\relbar\leftarrow}
\makeatother
\usepackage{amssymb}
\usepackage{amsthm}
\usepackage{amsmath}
\usepackage{centernot}

\newtheorem{intruleS}{Interpretation rule S\ignorespaces}
\newtheorem{intruleR}{Interpretation rule R\ignorespaces}

\usepackage{epstopdf}
\usepackage{authblk}
    
\usepackage{subcaption}
\usepackage{paralist}
\usepackage[numbers]{natbib}
\usepackage{url}
\usepackage{colortbl}
\usepackage{multirow}
\providecommand{\keywords}[1]{\textbf{\textit{Index Terms---}} #1}

\usepackage{stackengine}
\usepackage{scalerel}
\newcommand{\danger}{%
  \renewcommand\stacktype{L}%
  \scaleto{\stackon[0.8pt]{$\triangle$}{\tiny\bfseries !}}{1em}%
}

\usepackage{tikz}
\usetikzlibrary{arrows,positioning}
\tikzset{blob/.style={circle,draw,font=\sffamily\bfseries,minimum size=3.5em,text centered}}
\tikzset{g/.style={gray!35}}
\tikzset{t/.style={semithick}}
\tikzset{dt/.style={t,dashed}}

\begin{document}

\begin{textblock}{40}(1,1)
\noindent This accepted manuscript is licensed under the \href{http://creativecommons.org/licenses/by-nc-nd/4.0/}{CC BY-NC-ND 4.0 International License}.\\
The final version is published in \textit{NeuroImage}, 110:48-59, 2015, \href{http://dx.doi.org/10.1016/j.neuroimage.2015.01.036}{doi: 10.1016/j.neuroimage.2015.01.036}.
\end{textblock}

\title{Causal interpretation rules for encoding and decoding models in neuroimaging}

\date{}

\author[$^{1,*}$]{Sebastian Weichwald}\footnotetext[1]{Corresponding author.}
\author[$^1$]{Timm Meyer}
\author[$^2$]{Ozan Özdenizci}
\author[$^1$]{\\Bernhard Schölkopf}
\author[$^3$]{Tonio Ball}
\author[$^1$]{Moritz Grosse-Wentrup}

\affil[$^1$]{Max Planck Institute for Intelligent Systems, Tübingen, Germany\\ {\small\texttt{\{sweichwald, tmeyer, bs, moritzgw\}@tuebingen.mpg.de}}}
\affil[$^2$]{Sabancı University, Faculty of Engineering and Natural Sciences, Istanbul, Turkey\\ {\small\texttt{oozdenizci@sabanciuniv.edu}}}
\affil[$^3$]{Bernstein Center Freiburg, University of Freiburg, Freiburg, Germany\\ {\small\texttt{tonio.ball@uniklinik-freiburg.de}}}

\maketitle

\begin{abstract}
Causal terminology is often introduced in the interpretation of encoding and decoding models trained on neuroimaging data. In this article, we investigate which causal statements are warranted and which ones are not supported by empirical evidence. We argue that the distinction between encoding and decoding models is not sufficient for this purpose: relevant features in encoding and decoding models carry a different meaning in stimulus- and in response-based experimental paradigms. We show that only encoding models in the stimulus-based setting support unambiguous causal interpretations. By combining encoding and decoding models trained on the same data, however, we obtain insights into causal relations beyond those that are implied by each individual model type. We illustrate the empirical relevance of our theoretical findings on EEG data recorded during a visuo-motor learning task.
\end{abstract}


\newpage

\section{Introduction}

The question how neural activity gives rise to cognition is arguably one of the most interesting problems in neuroimaging \cite{hamann2001cognitive,ward2003synchronous,Atlas:2010}. Neuroimaging studies per se, however, only provide insights into neural correlates but not into neural causes of cognition \cite{ward2003synchronous,rees2002neural}. Nevertheless, causal terminology is often introduced in the interpretation of neuroimaging data. For instance, Hamann writes in a review on the neural mechanisms of emotional memory that ``Hippocampal activity in this study was correlated with amygdala activity, supporting the view that the amygdala \textit{enhances} explicit memory by \textit{modulating} activity in the hippocampus'' \cite{hamann2001cognitive}, and Myers et al.~note in a study on working memory that ``we tested [...]~whether pre-stimulus alpha oscillations measured with electroencephalography (EEG) \textit{influence} the encoding of items into working memory'' \cite{myers2014oscillatory} (our emphasis of causal terminology). The apparent contradiction between the prevalent use of causal terminology and the correlational nature of neuroimaging studies gives rise to the following question:  which causal statements are and which ones are not supported by empirical evidence?

We argue that the answer to this question depends on the experimental setting and on the type of model used in the analysis of neuroimaging data. Neuroimaging distinguishes between encoding and decoding models \cite{Naselaris2011}, known in machine learning as generative and discriminative models \cite{jordan2002discriminative}. Encoding models predict brain states, e.\,g.~BOLD activity measured by fMRI or event-related potentials measured by EEG/MEG, from experimental conditions \cite{Friston1994,Friston:2003,david2006dynamic}. Decoding models use machine learning algorithms to quantify the probability of an experimental condition given a brain state feature vector \cite{Mitchell2004,Pereira2009}. Several recent publications have addressed the interpretation of encoding and decoding models in neuroimaging, discussing topics such as potential confounds \cite{Todd2013,woolgar2014coping}, the dimensionality of the neural code \cite{davis2014differences}, and the relation of linear encoding and decoding models \cite{Haufe2014}. We contribute to this discussion by investigating, for each type of model, which causal statements are warranted and which ones are not supported by empirical evidence. Our investigation is based on the seminal work by Pearl \cite{Pearl2000} and Spirtes et al.~\cite{Spirtes2000} on causal inference (cf.~\cite{Ramsey:2010,Grosse-WentrupNeuroImage2011,waldorp2011effective,mumford2014bayesian} for applications of this framework in neuroimaging).
We find that the distinction between encoding and decoding models is not sufficient for this investigation. It is further necessary to consider whether models work in causal or anti-causal direction, i.\,e.~whether they model the effect of a cause or the cause of an effect \cite{Schoelkopf2012}. To accommodate this distinction, we distinguish between stimulus- and response-based paradigms. We then provide causal interpretation rules for each combination of experimental setting (stimulus- or response-based) and model type (encoding or decoding). We find that when considering one model at a time, only encoding models in stimulus-based experimental paradigms support unambiguous causal statements. Also, we demonstrate that by comparing encoding and decoding models trained on the same data, experimentally testable conditions can be identified that provide further insights into causal structure. These results enable us to reinterpret previous work on the relation of encoding and decoding models in a causal framework \cite{Todd2013,woolgar2014coping,Haufe2014}.

The empirical relevance of our theoretical results is illustrated on EEG data recorded during a visuo-motor learning task. We demonstrate that an encoding model allows us to determine EEG features that are effects of the instruction to rest or to plan a reaching movement, but does not enable us to distinguish between direct and indirect effects. By comparing relevant features in an encoding and a decoding model, we provide empirical evidence that sensorimotor $\mu$- and/or occipital $\alpha$-rhythms (8--14 Hz) are direct effects, while brain rhythms in higher cortical areas, including precuneus and anterior cingulate cortex, respond to the instruction to plan a reaching movement only as a result of the modulation by other cortical processes.

We note that while we have chosen to illustrate the empirical significance of our results on neuroimaging data, and specifically on EEG recordings, the provided causal interpretation rules apply to any encoding and decoding model trained on experimental data. This provides a guideline to researchers on how (not) to interpret encoding and decoding models when investigating the neural basis of cognition. A preliminary version of this work has been presented in \cite{Weichwald2014}.

\section{Methods}\label{sec:prelim}

We begin this section by introducing the causal framework by Pearl \cite{Pearl2000} and Spirtes et al.~\cite{Spirtes2000} that our work is based on (section \ref{sec:cbn}) and demonstrate how it leads to testable predictions for the causal statements cited in the introduction (section \ref{sec:causterm}).
We then introduce the distinction between causal and anti-causal encoding and decoding models (section \ref{sec:causalmodels}) and establish a connection between these models and causal inference (section \ref{sec:interrel}).
This connection enables us to present the causal interpretation rules for encoding and decoding models in section \ref{sec:rules}.
In section \ref{sec:jointinference}, we show that combining an encoding and a decoding model trained on the same data can provide further insights into causal structure.
We conclude this section by providing a reinterpretation of previous work on encoding and decoding models in a causal framework (section \ref{sec:previouswork}).

\subsection{Causal Bayesian Networks}\label{sec:cbn}
By $X$ we denote the finite set of $d$ random variables representing the brain state features, i.\,e.~$X = \{X_1,...,X_d\}$.
While these variables may correspond to any type of independent and identically distributed (iid) samples of $d$ brain state features, it is helpful to consider bandpower features of different EEG channels, trial-averaged BOLD activity at various cortical locations, or mean spike rates of multiple neurons as possible examples.
By $C$ we denote the random variable representing the (usually discrete) experimental condition. $C$ stands for a stimulus ($C\equiv S$) or response ($C\equiv R$) variable and it will be made clear when $C$ is restricted to either particular case.
For convenience, we denote the set of all random variables by $\widehat{X}=\{C,X_1,...,X_d\}$.
Throughout this article, we denote marginal, conditional and joint distributions by $P(X)$, $P(X|C)$ and $P(X,C)$, respectively (overloading the notation of $P$).
For our theoretical investigations, we assume that the involved distributions have probability mass or density functions (PMFs or PDFs) with values denoted by $P(x), P(x|c)$ and $P(x,c)$ respectively, again overloading the notation of $P$ while it is always clear from the argument which function is meant.
We use the common notations for independence and conditional independence
\begin{align*}
    &X\indep C &&:\!\iff P(X|C)=P(X), \\
    &X\indep C | Y &&:\!\iff P(X|C,Y)=P(X|Y).
\end{align*}

In the framework of Causal Bayesian Networks (CBNs) \cite{Pearl2000,Spirtes2000}, a variable $X_i$ is said to be a cause of another variable $X_j$ if the distributions $P(X_j | \text{do}(X_i = x_i))$ are sensitive to $x_i$ (cf.~\cite{Pearl2000}, p.~24f.). In this notation, the intervention \mbox{$\text{do}(X_i = x_i)$} signifies that $X_i$ is externally set to a constant $x_i$, possibly resulting in a change of the distribution of $X_j$. The framework of CBNs thus defines cause-effect relations in terms of the impact of external manipulations. This is in contrast to frameworks that define causality in terms of information transfer \cite{Granger:1969, Roebroeck:2005, Lizier:2010}.

Causal relations between variables in CBNs are represented by directed acyclic graphs (DAGs).
If we find a directed edge $X_i \to X_j$, we call $X_i$ a direct cause of $X_j$ and $X_j$ a direct effect of $X_i$.
In case there is no directed edge but at least one directed path $X_i \dashrightarrow X_j$, we call $X_i$ an indirect cause of $X_j$ and $X_j$ an indirect effect of $X_i$.
Note that the terms \emph{(in-)direct cause/effect} depend on the set $\widehat{X}$ of observed variables: consider $\widehat{X} = \{C,X_1,X_2\}$ and the causal DAG $C \to X_1 \to X_2$. Then $C \dashrightarrow X_2$ and $C \centernot\to X_2$ wrt.~$\widehat{X}$, while $C \to X_2$ wrt.~$\widehat{X}'=\{C,X_2\}$.
That is, whether a cause or effect is direct or indirect depends on the set of observed brain state features. We omit the considered set of nodes if it is clear from the context.

To establish a link between conditional independences and DAGs the following concepts are required:
\begin{description}
    \item[$d$-separation]
        Disjoint sets of nodes $A$ and $B$ are d-separated by another disjoint set of nodes $C$ if and only if all $a \in A$ and $b \in B$ are d-separated by $C$.
        Two nodes $a \neq b$ are d-separated by $C$ if and only if every path between $a$ and $b$ is blocked by $C$.
        A path between nodes $a$ and $b$ is blocked by $C$ if and only if there is an intermediate node $z$ on the path such that (i) $z \in C$ and $z$ is a tail-to-tail ($\gets z \to$) or head-to-tail ($\to z \to$) or (ii) $z$ is head-to-head ($\to z \gets$) and neither $z$ nor any of its descendants is in $C$.
\item[Causal Markov Condition (CMC)]
        The CMC expresses the notion that each node in a causal DAG becomes independent of its non-descendants given its direct causes, i.\,e. that the causal structure implies certain (conditional) independences.
    \item[Faithfulness]
        The faithfulness assumption states that all (conditional) independences between the random variables of a DAG are implied by its causal structure, i.\,e. there are no more (conditional) independences than those implied by the CMC.
\end{description}

Assuming faithfulness and the causal Markov condition, d-separation is equivalent to conditional independence, i.\,e.~$C$ d-separates $A$ and $B$ if and only if $A$ and $B$ are independent given $C$ \cite{Spirtes2000}.
The following three examples are the most relevant instances of d-separation for our following arguments.
Firstly, consider the chain $X_0 \to X_1 \to X_2$. Here, $X_1$ d-separates $X_0$ and $X_2$ by blocking the directed path from $X_0$ to $X_2$. This implies that $X_0 \indep X_2 | X_1$. Secondly, consider the fork $X_0 \leftarrow X_1 \to X_2$. Here, $X_1$ d-separates $X_0$ and $X_2$, as $X_1$ is a joint cause of $X_0$ and $X_2$. This again implies that $X_0 \indep X_2 | X_1$. Thirdly, consider the collider $X_0 \to X_1 \leftarrow X_2$. In this case, $X_1$ does not d-separate $X_0$ and $X_2$. As $X_1$ is a joint effect of $X_0$ and $X_2$, it unblocks the previously blocked path between $X_0$ and $X_2$, implying that $X_0 \not\indep X_2 | X_1$.

The equivalence between d-separation and conditional independence enables us to infer causal relations between variables in $\widehat{X}$ from observational data. By identifying conditional independences that hold in our data, and mapping them onto the equivalent d-separations, we gain knowledge about the causal structures that can give rise to our data. This link forms the basis of the inference rules we describe in section \ref{sec:rules} and \ref{sec:jointinference}. We refer the interested reader to \cite{mumford2014bayesian} for a  more exhaustive introduction to this causal inference framework in the context of neuroimaging.

\subsection{Causal terminology and CBNs}\label{sec:causterm}

We now demonstrate how the causal statements, that we cited in the introduction to motivate our work, can be expressed in the framework of CBNs.

Firstly, consider the statement ``Hippocampal activity in this study was correlated with amygdala activity, supporting the view that the amygdala enhances explicit memory by modulating activity in the hippocampus'' \cite{hamann2001cognitive}. Here, it is implicitly assumed that hippocampal activity is a cause of explicit memory; that is, manipulating hippocampal activity results in measurable changes in explicit memory. In the notation of CBNs, this is expressed as \textit{hippocampal activity} $\to$ \textit{explicit memory}. Further, it is implied that the amygdala enhances explicit memory via modulating activity in the hippocampus; that is, manipulating activity in the amygdala leads to measurable changes in hippocampal activity, which results in changes in explicit memory. This gives the causal hypothesis \textit{amygdala activity} $\to$ \textit{hippocampal activity} $\to$ \textit{explicit memory}. Assuming faithfulness and the CMC, this causal hypothesis makes the empirically testable predictions \textit{amygdala activity} $\centernot\indep $ \textit{explicit memory} and \textit{amygdala activity} $\indep$ \textit{explicit memory} $|$ \textit{hippocampal activity}.

Secondly, consider the statement ``we tested [...]~whether pre-stimulus alpha oscillations measured with electroencephalography (EEG) influence the encoding of items into working memory'' \cite{myers2014oscillatory}. The authors thereby express the notion that pre-stimulus alpha oscillations are a cause of working memory; that is, manipulating pre-stimulus alpha oscillations has a measurable effect on working memory. This gives the causal hypothesis \textit{pre-stimulus alpha oscillations} $\to$ \textit{working memory}. Again assuming faithfulness and the CMC, this causal hypothesis makes the empirically testable prediction \textit{pre-stimulus alpha oscillations} $\centernot\indep$ \textit{working memory}.

In the following, we investigate how these empirical predictions can be tested by encoding and decoding models, and under which conditions these predictions are sufficient to prove a causal hypothesis. In section \ref{sec:revisit} we revisit those two examples illustrating our theoretical results.

\subsection{Causal and anti-causal encoding and decoding models}\label{sec:causalmodels}
An encoding model $P(X|C)$ represents how various experimental conditions are encoded in the brain state feature vector. We ask ``How does the brain state look like given a certain experimental condition?''. Examples for encoding models are the general linear model \cite{Friston1994} or the class-conditional mean $E\{X|C\}$.

A decoding model $P(C|X)$ represents how experimental conditions can be inferred from the brain state feature vector \cite{Mitchell2004}. We ask ``How likely is an experimental condition given a certain brain state feature vector?''.
Decoding models are for example obtained using support vector machines (SVMs) or linear regression with $X$ as the independent and $C$ as the dependent variable.

A priori, encoding and decoding models are oblivious to the causal relation between experimental conditions and brain state features, i.\,e.~they disregard whether they model the effect of a cause or the cause of an effect \cite{Schoelkopf2012}. As we show in section \ref{sec:interrel}, however, this has implications for their interpretation. We hence introduce the distinction between stimulus- and response-based experimental paradigms.

We categorize an experimental setup as stimulus-based, if the experimental conditions precede the measured brain states. An example of a stimulus-based setup is the investigation of the brain's activity when exposed to auditory stimuli. As the auditory stimuli precede the measured brain activity, the brain state features cannot be a cause of the stimuli. In the stimulus-based setting, an encoding model $P(X|C)\equiv P(X|S)$ works in causal direction, as it models the effect of a cause. A decoding model $P(C|X)\equiv P(S|X)$ works in anti-causal direction, as it models the cause of an effect. We note that in the stimulus-based setting we can control for and randomize the stimulus, i.\,e.~we can externally set the stimulus to a desired value denoted by $\text{do}(S = s)$.

We categorize an experimental setup as response-based, if the measured brain states precede the experimental conditions. An example of a response-based setup is the prediction of the laterality of a movement from pre-movement brain state features. As the measured brain state features precede the actual movement, the movement cannot be a cause of this brain activity. In the response-based setting, an encoding model $P(X|C)\equiv P(X|R)$ works in anti-causal direction, as it models the cause of an effect. A decoding model $P(C|X)\equiv P(R|X)$ works in causal direction, as it models the effect of a cause. We note that in contrast to the stimulus-based setting, we cannot control for and randomize the response, i.\,e.~we cannot set the response to a desired value by an external intervention.

We also note that more complex experimental paradigms, in which responses act again as stimuli \cite{Gomez2011}, can also be categorized in this way by considering time-resolved variables, e.\,g.~stimulus$[t_0]$ $\rightarrow$ brain activity$[t_1]$ $\rightarrow$ response$[t_2] \rightarrow$ stimulus$[t_3]$.

In the following, we hence distinguish between four types of models:
\begin{enumerate}
 \item Causal encoding models -- $P(X|S)$
 \item Anti-causal decoding models -- $P(S|X)$
 \item Anti-causal encoding models -- $P(X|R)$
 \item Causal decoding models -- $P(R|X)$
\end{enumerate}

\subsection{Relating encoding  and decoding models to causal inference}\label{sec:interrel}

In this section, we establish a link between causal inference and the identification of relevant features in encoding and decoding models. This link forms the basis for the causal interpretation rules in the next section.

When using an encoding model to analyze neuroimaging data, we wish to identify features that show a statistically significant variation across experimental conditions. In practice, this can be carried out by a variety of methods including but not limited to a general linear model \cite{Friston1994}, class-conditional differences in mean activation \cite{Delorme:2004}, and non-linear independence tests \cite{gretton2008kernel,Grosse-WentrupNeuroImage2011}. Common to all these approaches is that they admit univariate statistical tests to quantify the likelihood of the data under the null-hypothesis $X_i \indep C$. Features, for which the null-hypothesis of independence is rejected, are considered relevant for the encoding model in this experimental paradigm. Features with insufficient evidence for rejection of the null-hypothesis are considered irrelevant for the present encoding model. The set of relevant and irrelevant features in an encoding model is subsequently denoted as $X^{\text{+enc}}$ and $X^{\text{--enc}}$, respectively. Assuming faithfulness, the relevance of features in an encoding model can be translated into d-separation properties that provide insights into the causal structure of the data-generating process. In particular, features in $X^{\text{--enc}}$ are d-separated from the experimental condition by the empty set while features in $X^{\text{+enc}}$ are not d-separated from the experimental condition by the empty set.

When using a decoding model to analyze neuroimaging data, we wish to identify features that help in decoding the experimental condition; that is, we wish to determine for each feature if its removal increases the minimum Bayes error. If $X_1 \indep C | X \setminus X_1$ the Bayes classifier $\operatorname{argmax}_{c} P(c|x_1,...,x_d) = \operatorname{argmax}_{c} P(c|x_2,...,x_d)$ and hence also the Bayes error rate remain unchanged. This leads to the mathematical concept of conditional independence for a feature's relevance in decoding \cite{strobl2008conditional}: a feature $X_i$ is relevant for decoding if $X_i {\not\indep} C | X \setminus X_i$. In practice, one commonly used approach is to permute a feature with respect to the experimental conditions and check if this results in a statistically significant decrease in decoding accuracy \cite{breiman2001random}.
Features, for which the null-hypothesis of conditional independence is rejected, are considered relevant for the decoding model. Features with insufficient evidence for rejection of the null-hypothesis are considered irrelevant for decoding the experimental condition given the other features. The set of relevant and irrelevant features in a decoding model is subsequently denoted as $X_{\text{+dec}}$ and $X_{\text{--dec}}$, respectively.
Assuming faithfulness again, every feature $X_i \in X_{\text{--dec}}$ is d-separated from the experimental condition by the set $X \setminus X_i$, while every feature $X_i \in X_{\text{+dec}}$ is not d-separated from the experimental condition by the set $X \setminus X_i$.

The link between the relevance of features in encoding and decoding models and d-separation properties allows us to provide the causal interpretation rules given in the next section. For our theoretical arguments, we assume that we can identify all relevant features for each type of model, that is, we assume we can identify all (conditional) (in-)dependence relations. We discuss the practical intricacies of identifying independence relations on finite empirical data in section \ref{sec:disc}.

\subsection{Causal interpretation rules for encoding and decoding models}\label{sec:rules}

\subsubsection{Causal encoding models $P(X|S)$}\label{sec:intA}
According to Reichenbach's principle \cite{Reichenbach1956}, the dependency between $S$ and $X_i \in X^\text{+enc}$ implies that $S \rightarrow X_i$, $S \leftarrow X_i$, or $S \leftarrow H \rightarrow X_i$ with $H$ a joint common cause of $S$ and $X_i$. In the stimulus-based setting, we can control for and randomize the stimulus. This enables us to rule out the last two cases and conclude that $S \rightarrow X_i$, i.\,e.~the features in $X^\text{+enc}$ are effects of $S$ \cite{Holland1986}.

In contrast, for $X_i \in X^\text{--enc}$ we have $S \indep X_i$, which allows us to conclude that features in $X^\text{--enc}$ are not effects of $S$.

As such, all relevant features in a causal encoding model are effects of $S$, while irrelevant features are not effects of $S$. We hence have the following two interpretation rules:
\begin{intruleS} For $C\equiv  S$:
    \begin{align*}
        X_i \in X^\text{+enc} \iff &X_i \text{ is an effect of } S \text{, i.\,e.~} S \dashrightarrow X_i
    \end{align*}
\end{intruleS}

\begin{intruleS} For $C\equiv  S$:
    \begin{align*}
        X_i \in X^\text{--enc} \iff &X_i \text{ is not an effect of } S \text{, i.\,e.~} S \centernot\dashrightarrow X_i
    \end{align*}
\end{intruleS}

\subsubsection{Anti-causal decoding models $P(S|X)$}\label{sec:antidec}
We describe two counterexamples that show that features in $X_\text{+dec}$ are not necessarily effects of $S$ and that features in $X_\text{--dec}$ can be effects of $S$.

Firstly, assume $S \to X_1 \gets X_2$. Since $X_1$ does not d-separate $S$ and $X_2$, i.\,e.~$S \not\indep X_2|X_1$, we obtain that $X_2 \in X_\text{+dec}$ although $X_2$ is not an effect of $S$.

Secondly, assume $S \to X_1 \to X_2$. Since $X_1$ d-separates $S$ and $X_2$, i.\,e.~$S \indep X_2 | X_1$, we have $X_2 \in X_\text{--dec}$ although $X_2$ is an effect of $S$.

This establishes that interpreting relevant features in anti-causal decoding models has two drawbacks.
Firstly, features in $X_\text{+dec}$ are only potentially effects of $S$.
And secondly, effects of $S$ might be missed, since effects are not necessarily relevant for decoding the experimental condition. This yields the following two interpretation rules:

\begin{intruleS} For $C\equiv  S$:
    \begin{align*}
        X_i \in X_\text{+dec} \centernot\iff &X_i \text{ is an effect of } S \text{, i.\,e.~} S \dashrightarrow X_i
    \end{align*}
\end{intruleS}

\begin{intruleS} For $C\equiv  S$:
    \begin{align*}
        X_i \in X_\text{--dec} \centernot\iff &X_i \text{ is not an effect of } S \text{, i.\,e.~} S \centernot\dashrightarrow X_i
    \end{align*}
\end{intruleS}

\subsubsection{Anti-causal encoding models $P(X|R)$}\label{sec:antienc}
According to Reichenbach's principle, the dependency between $X_i \in X^{\text{+enc}}$ and $R$ implies that $X_i \rightarrow R$, $X_i \leftarrow R$, or $X_i \leftarrow H \rightarrow R$ with $H$ a joint common cause of $X_i$ and $R$.
A priori, we know that brain activity $\to$ response. This enables us to rule out the case $X_i \gets R$.
We now show that one cannot uniquely determine which of the last two scenarios is the case, i.\,e.~features in $X^{\text{+enc}}$ are only potentially causes of $R$.
Consider $X_2 \gets X_1 \to R$. In this case, we have $X_1 \not\indep R$ and $X_2 \not\indep R$ and therefore $X_1,X_2 \in X^{\text{+enc}}$. But note that $X_1 \to R$ while $X_2 \centernot\dashrightarrow R$, i.\,e.~$X_2$ is not a cause of $R$. This shows that features in $X^\text{+enc}$ are not necessarily causes of $R$.

Features in $X^\text{--enc}$, on the other hand, are independent of $R$ and hence cannot be causes of $R$.

As such, not all relevant features in anti-causal encoding models are causes of $R$, while irrelevant features are indeed not causal for $R$. We hence have the following two interpretation rules:
\begin{intruleR} For $C\equiv  R$:
    \begin{align*}
        X_i \in X^\text{+enc} \iff &X_i \text{ is only potentially a cause of } R, \\
        &\text{i.\,e.~} X_i \dashrightarrow R \text{ or } X_i \dashleftarrow H \dashrightarrow R
    \end{align*}
\end{intruleR}

\begin{intruleR} For $C\equiv  R$:
    \begin{align*}
        X_i \in X^\text{--enc} \implies &X_i \text{ is not a cause of } R,\\
        &\text{i.\,e.~} X_i \centernot\dashrightarrow R
    \end{align*}
\end{intruleR}

\subsubsection{Causal decoding models $P(R|X)$}\label{sec:intD}
We describe two counterexamples that show that one can neither conclude that features in $X_\text{+dec}$ are causes of $R$ nor that features in $X_\text{--dec}$ are not causes of $R$.

Firstly, consider $X_2 \to X_1 \gets H \to R$ where $H$ is a hidden common cause of $X_1$ and $R$, yet $H$ is a non-observable brain state feature.
We have $X_1,X_2 \in X_\text{+dec}$ in this example, as $X_1$ as well as $X_2$ are d-separated from $R$ only by the non-observable common cause $H$. But both $X_1$ and $X_2$ are not causes of $R$.

Secondly, assume $X_2 \to X_1 \to R$. Since $X_1$ d-separates $X_2$ and $R$, i.\,e.~$X_2 \indep R | X_1$, we have $X_2 \in X_\text{--dec}$ although $X_2$ is a cause of $R$.

This establishes that interpreting relevant features in causal decoding models has two drawbacks. Firstly, features in $X_\text{+dec}$ are not necessarily causes of $R$.
Secondly, causes of $R$ might be missed, since causes are not necessarily relevant for decoding the experimental condition. This yields the following two interpretation rules:

\begin{intruleR} For $C\equiv  R$:
    \begin{align*}
        X_i \in X_\text{+dec} &\centernot\iff X_i \text{ is a cause of } R \text{, i.\,e.~} X_i \dashrightarrow R
    \end{align*}
\end{intruleR}

\begin{intruleR} For $C\equiv  R$:
    \begin{align*}
        X_i \in X_\text{--dec} &\centernot\iff X_i \text{ is not a cause of } R \text{, i.\,e.~} X_i \centernot\dashrightarrow R
    \end{align*}
\end{intruleR}

\subsubsection{Subsumption}\label{sec:intE}
In the previous subsections, we showed that the interpretation of relevant features in encoding and decoding models depends on the experimental setting. This justifies our argument that the distinction of encoding and decoding models is not sufficient to determine their causal interpretation. In particular we argued that, without employing further assumptions,

\begin{enumerate}[1. (cf. \thesubsection.1.)]
        \item causal encoding models $P(X|S)$ identify \emph{all} effects of $S$ in $\widehat{X}$.
        \item anti-causal decoding models $P(S|X)$ only identify \emph{some} features being \emph{potentially} effects of $S$ in $\widehat{X}$.
        \item anti-causal encoding models $P(X|R)$ identify \emph{all} features being \emph{potentially} causes of $R$ in $\widehat{X}$.
        \item causal decoding models $P(R|X)$ only identify \emph{some} features being \emph{potentially} causes of $R$ in $\widehat{X}$.
\end{enumerate}

\subsection{Causal interpretation rules for combined encoding and decoding models}\label{sec:jointinference}

In section \ref{sec:rules}, we showed that only encoding models in a stimulus-based setting support unambiguous causal statements. This result appears to imply that decoding models, despite their gaining popularity in neuroimaging, are of little value for investigating the neural causes of cognition. In the following, we argue that this is not the case. Specifically, we show that by combining encoding and decoding models, we gain insights into causal structure that are not possible by investigating each type of model individually. In analogy to section \ref{sec:rules}, we again distinguish between stimulus- and response-based paradigms.

For a combined analysis of encoding and decoding models, we intuitively extend our notation and define the following four sets of brain state features, which yield a disjoint partition of $X$:
\begin{alignat*}{4}
    &X^\text{+enc}_\text{+dec} &&:= X^\text{+enc} &&\cap X_\text{+dec}\\
    &X^\text{+enc}_\text{--dec} &&:= X^\text{+enc} &&\cap X_\text{--dec}\\
    &X^\text{--enc}_\text{+dec} &&:= X^\text{--enc} &&\cap X_\text{+dec}\\
    &X^\text{--enc}_\text{--dec} &&:= X^\text{--enc} &&\cap X_\text{--dec}
\end{alignat*}
We now provide causal interpretation rules for each type of feature set and experimental setting.

\subsubsection{Stimulus-based setting}

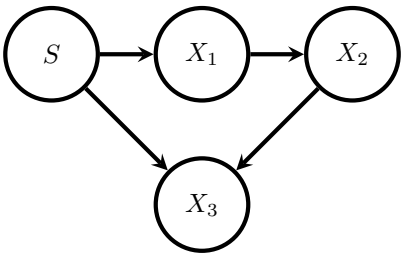
\begin{figure}[t]
\centering
\begin{tikzpicture}[->,>=stealth,auto,on grid,ultra thick]
    \node[blob](a) at(0,2) {$S$};
    \node[blob](h) at(2,2) {$X_1$};
    \node[blob](b) at(4,2) {$X_2$};
    \node[blob](R) at(2,0) {$X_3$};

  \path[every node/.style={font=\sffamily\small}]
    (a) edge node [left] {} (h)
    (h) edge node [left] {} (b)
    (a) edge node [left] {} (R)
    (b) edge node [left] {} (R);
\end{tikzpicture}
\caption{The causal DAG illustrates that $X_2 \in X^\text{+enc}_\text{+dec}$ does not imply that $X_2$ is a direct effect of $S$ wrt. $\widehat{X}$.}\label{fig:direffect}
\end{figure}

\paragraph{Features relevant in encoding and relevant in decoding: $X^\text{+enc}_\text{+dec}$}\ \\
As $X_i \in X^\text{+enc}$, it is an effect of $S$ (rule S1). Intuitively, the fact that furthermore $X_i \in X_\text{+dec}$ suggests that $X_i$ is in some sense closer to $S$. Figure \ref{fig:direffect}, however, establishes that this intuition is not correct. Since $S \not\indep X_2$ and $S \not\indep X_2 | \{X_1,X_3\}$, we have $X_2 \in X^\text{+enc}_\text{+dec}$ even though $X_2$ is an indirect effect of $S$. Hence, features that are relevant in both encoding and decoding do not provide further insights into causal structure. This leads to the following interpretation rule (note the missing bi-implication compared to interpretation rule S1):

\begin{intruleS} For $C\equiv  S$:
    \begin{align*}
        X_i \in X^\text{+enc}_\text{+dec} \implies &X_i \text{ is an effect of } S \text{, i.\,e.~} S \dashrightarrow X_i
    \end{align*}
\end{intruleS}

\paragraph{Features relevant in encoding and irrelevant in decoding: $X^\text{+enc}_\text{--dec}$}\ \\
As $X_i \in X^\text{+enc}$, it holds that $X_i$ is an effect of $S$ (rule S1). Only looking at the encoding side, however, does not determine whether $X_i$ is a direct ($S \to X_i$) or indirect effect ($S \dashrightarrow X_i$) of $S$ wrt. $\widehat{X}$. As $X_i \in X_\text{--dec}$, however, $S$ and $X_i$ are d-separated by the set $X \setminus X_i$ (cf.~section \ref{sec:interrel}). This rules out that $X_i$ is a direct effect of $S$ wrt. $\widehat{X}$, leaving $S \dashrightarrow X_i$ as the only explanation. This leads to the following interpretation rule:

\begin{intruleS} For $C\equiv  S$:
    \begin{align*}
        X_i \in X^\text{+enc}_\text{--dec} \implies &X_i \text{ is an indirect effect of } S \text{ wrt. } \widehat{X}
    \end{align*}
\end{intruleS}

\paragraph{Features irrelevant in encoding and relevant in decoding: $X^\text{--enc}_\text{+dec}$}\ \\
As $X_i \in X^\text{--enc}$, it is not an effect of $S$ (rule S2). As further $X_i \in X_\text{+dec}$, $X_i$ and $S$ are not d-separated by $X\setminus X_i$. This implies that $X_i$ is either a cause of variables in $X\setminus X_i$ or that $X_i$ and $X\setminus X_i$ share at least one common cause. Examples of these scenarios are $S \to X_1 \gets X_2$ and $S \to X_1 \gets H \to X_2$. In both cases, knowledge about $X_2$ can be used to better decode $S$ from $X_1$ by removing variations in $X_1$ that are not due to $S$. This leads to the following interpretation rule:

\begin{intruleS} For $C\equiv  S$:
    \begin{align*}
        X_i \in X^\text{--enc}_\text{+dec} \implies &X_i \text{ provides brain state context wrt. } S
    \end{align*}
\end{intruleS}

\paragraph{Features irrelevant in encoding and irrelevant in decoding: $X^\text{--enc}_\text{--dec}$}\ \\
Features in $X^\text{--enc}_\text{--dec}$ are neither effects of $S$ (rule S2) nor do they provide brain state context wrt. $S$. Hence, they can be considered irrelevant for the present experimental context. This is summarized in the following interpretation rule:

\begin{intruleS} For $C\equiv  S$:
    \begin{align*}
        X_i \in X^\text{--enc}_\text{--dec} \implies &X_i \text{ is neither an effect of $S$ nor provides brain state context}
    \end{align*}
\end{intruleS}

\subsubsection{Response-based setting}

\paragraph{Features relevant in encoding and relevant in decoding: $X^\text{+enc}_\text{+dec}$}\ \\
As $X_i \in X^\text{+enc}$, it is potentially a cause of $R$ (rule R1).
Intuitively, the fact that furthermore $X_i \in X_\text{+dec}$ suggests that $X_i$ is in some sense closer to $R$, e.\,g. that $X_i$ is a cause of $R$.
The DAG $X_1 \gets H \to R$, however, establishes that this intuition is not correct. Since $R \not\indep X_1$, we have $X_1 \in X^\text{+enc}_\text{+dec}$, even though $X_1$ is not a cause of $R$. Hence, features that are relevant in both encoding and decoding do not provide further insights into causal structure. This leads to the following interpretation rule (note the missing bi-implication compared to interpretation rule R1):

\begin{intruleR} For $C\equiv  R$:
    \begin{align*}
        X_i \in X^\text{+enc}_\text{+dec} \implies &X_i \text{ is only potentially a cause of } R, \\
        &\text{i.\,e.~} X_i \dashrightarrow R \text{ or } X_i \dashleftarrow H \dashrightarrow R
    \end{align*}
\end{intruleR}

\paragraph{Features relevant in encoding and irrelevant in decoding: $X^\text{+enc}_\text{--dec}$}\ \\
As $X_i \in X^\text{+enc}$, it holds that $X_i \dashrightarrow R$ or $X_i \dashleftarrow H \dashrightarrow R$ (rule R1).
Although we cannot distinguish between those two scenarios, we can learn more about $X_i$ from its irrelevance in decoding. $X_i \to R$ would imply that $X_i$ and $R$ are not d-separated by $X\setminus X_i$ and hence $X_i \in X^\text{+enc}_\text{+dec}$. Thus, $X_i$ cannot be a direct cause of $R$ wrt. $\widehat{X}$. This leads to the following interpretation rule:

\begin{intruleR} For $C\equiv  R$:
    \begin{align*}
        X_i \in X^\text{+enc}_\text{--dec} \implies &X_i \text{ is no direct cause of } R \text{ wrt. } \widehat{X}
    \end{align*}
\end{intruleR}

\paragraph{Features irrelevant in encoding and relevant in decoding: $X^\text{--enc}_\text{+dec}$}\ \\
As $X_i \not\in X^\text{--enc}$, it is not a cause of $R$ (rule R2). As further $X_i \in X_\text{+dec}$, $X_i$ and $R$ are not d-separated by $X\setminus X_i$. This implies that $X_i$ is either a cause of variables in $X\setminus X_i$ or that $X_i$ and $X\setminus X_i$ share at least one common cause. An example of this scenario is $X_2 \to X_1 \gets H \to R$. Here, knowledge about $X_2$ can be used to better decode $R$ from $X_1$ by removing variations in $X_1$ that are not due to $H$. In analogy to the stimulus-based setting, this leads to the following interpretation rule:

\begin{intruleR} For $C\equiv  R$:
    \begin{align*}
        X_i \in X^\text{--enc}_\text{+dec} \implies &X_i \text{ provides brain state context wrt. } R
    \end{align*}
\end{intruleR}

\paragraph{Features irrelevant in encoding and irrelevant in decoding: $X^\text{--enc}_\text{--dec}$}\ \\
Features in $X^\text{--enc}_\text{--dec}$ are neither causes of $R$ (rule R2) nor do they provide brain state context wrt. $R$. Hence, they can be considered irrelevant for the present experimental context. This leads to the following interpretation rule:

\begin{intruleR} For $C\equiv  S$:
    \begin{align*}
        X_i \in X^\text{--enc}_\text{--dec} \implies &X_i \text{ is neither a cause of $R$ nor provides brain state context}
    \end{align*}
\end{intruleR}

\subsubsection{Subsumption}

We summarize all causal interpretation rules in Table~\ref{tab:intrule}.
Combining encoding and decoding models is particularly useful if features turn out to be relevant in only one type of model: features only relevant in encoding are not direct effects/causes, while features only relevant for decoding do provide brain state context wrt. $S/R$ while not being effects/causes.

We further note that due to the possibility of hidden confounders, features potentially being a cause and genuine causes cannot be distinguished in the response-based setting without introducing further assumptions.

\begin{table*}[h]
    \small
    \centering
    \renewcommand{\arraystretch}{1.3}
    \caption{Causal interpretation rules for relevant ($\surd$) and/or irrelevant ($\times$) features $X_i$ in encoding and decoding models for stimulus- ($C\equiv S$) and response-based ($C\equiv R$) paradigms.}
    \centering
    \tabcolsep=0.15cm
\begin{adjustbox}{center}
\begin{tabular}{l|c|c|l|c}
    & relevance in encoding & relevance in decoding & causal interpretation & rule \\

    \hline
    \hline

    \parbox[t]{2mm}{\multirow{8}{*}{\rotatebox[origin=c]{90}{stimulus-based}}}
    & $\surd$ & \cellcolor{gray}        & $X$ effect of $S$ & S1\\
    \cline{2-5}
    & $\times$ & \cellcolor{gray}       & $X$ no effect of $S$ & S2\\
    \cline{2-5}
    & \cellcolor{gray} & $\surd$        & \danger\ inconclusive & S3\\
    \cline{2-5}
    & \cellcolor{gray} & $\times$       & \danger\ inconclusive & S4\\
    \cline{2-5}
    & $\surd$ & $\surd$                 & $X$ effect of $S$ & S5\\
    \cline{2-5}
    & $\surd$ & $\times$                & $X$ indirect effect of $S$ & S6\\
    \cline{2-5}
    & $\times$ & $\surd$                & provides brain state context & S7\\
    \cline{2-5}
    & $\times$ & $\times$               & neither effect nor provides brain state context & S8\\

    \hline
    \hline

    \parbox[t]{2mm}{\multirow{8}{*}{\rotatebox[origin=c]{90}{response-based}}}
    & $\surd$ & \cellcolor{gray}        & \danger\ inconclusive &R1\\
    \cline{2-5}
    & $\times$ & \cellcolor{gray}       & $X$ no cause of $R$ & R2\\
    \cline{2-5}
    & \cellcolor{gray} & $\surd$        & \danger\ inconclusive & R3\\
    \cline{2-5}
    & \cellcolor{gray} & $\times$       & \danger\ inconclusive & R4\\
    \cline{2-5}
    & $\surd$ & $\surd$                 & \danger\ inconclusive & R5\\
    \cline{2-5}
    & $\surd$ & $\times$                & $X$ no direct cause of $R$ & R6\\
    \cline{2-5}
    & $\times$ & $\surd$                & provides brain state context & R7\\
    \cline{2-5}
    & $\times$ & $\times$               & neither cause nor provides brain state context & R8\\

    \hline
\end{tabular}\label{tab:intrule}
\end{adjustbox}
\end{table*}

\subsection{Reinterpretation of previous work in a causal framework}\label{sec:previouswork}

In this section, we discuss previous work on the interpretation of encoding and decoding models in light of the causal interpretation rules that we introduced in the previous sections.

\subsubsection{Potential confounds}

In \cite{Todd2013,woolgar2014coping} the problem of potential confounds in multi-voxel pattern analysis (MVPA) has been discussed.
In particular, Todd et al.~demonstrated that decoding models may determine brain state features as relevant that are statistically independent of the experimental condition.
This finding is confirmed by interpretation rules S7 and R7 that we presented in section~\ref{sec:jointinference}.
These rules reveal that what Todd et al.~termed confounds are exactly those features that are irrelevant in encoding and relevant in decoding.
Hence, our reinterpretation in a causal framework re-emphasizes the potential problem highlighted in \cite{Todd2013} and additionally allows to specify the characteristics of such features, i.\,e.~being irrelevant in encoding and relevant in decoding.

In contrast to Todd et al.~we do not term those features confounds, as this terminology suggests that such features invalidate interpretation of other features and cannot be interpreted. Instead, we propose to use the terminology \textit{features that provide brain state context}. Knowledge about the specific characteristics of features that provide brain state context wrt. $S/R$ can indeed lead to interesting causal statements, as demonstrated in \cite{Grosse-WentrupNeuroImage2011}: under the additional assumption of causal sufficiency, the assumption that all causally relevant features have been observed, a causal influence of gamma oscillations ($\gamma$) on the sensorimotor rhythm was concluded from the fact that $\gamma \in X^\text{--enc}_\text{+dec}$.

As we have shown in the previous section, encoding and decoding models provide complementary information.
We hence argue that the problem discussed in \cite{Todd2013,woolgar2014coping} is not a shortcoming of decoding models, but rather a useful feature. Decoding models provide insights into causal structure that cannot be gained by only investigating encoding models. Pitfalls are not due to weaknesses of decoding models, but a result of negligent interpretation of relevant features.
Our findings are thus in line with Todd et al.~and Woolgar et al.~and strengthen the point that MVPA results need to be interpreted carefully.

\subsubsection{Linear encoding and decoding models}

In \cite{Haufe2014} Haufe et al.~demonstrated that linear backward models, i.\,e.~linear decoding models, ``cannot be interpreted in terms of the studied brain processes'', as a large weight does not necessarily correspond to a feature that picks up the signal and, vice-versa, a feature that picks up the signal does not necessarily have a large weight.
The causal interpretation rules S3, S4 and R3, R4 extend this argument to non-linear decoding models and yield a reinterpretation of these findings in the framework of CBNs.
What is more, our distinction of stimulus- and response-based experimental settings allows us to specify the finding that linear forward models, i.\,e.~linear encoding models, are not affected by this problem: in accordance with Haufe et al.~we showed that in the stimulus-based setting encoding models, both linear and non-linear, allow unambiguous causal statements. However, in the response-based setting only those features irrelevant in encoding are unproblematic to be interpreted as non-causes of $R$, while the meaning of relevant features remains ambiguous in this setting.

For the linear case Haufe et al.~presented an intriguing way to obtain an encoding from a decoding model.
Linear models only allow to test for correlation, and jointly Gaussian random variables are the only general case for which the lack of correlation implies independence.
As such, if one is willing to assume that all variables are jointly Gaussian, this method together with our findings yields an easy way to enrich causal interpretation when using linear models.
Firstly, in contrast to the decoding model one started with, the encoding model obtained by inversion allows for causal statements on its own (interpretation rules S1, S2 or R2).
Secondly, since then both an encoding and a decoding model are at hand, also interpretation rules S5-8 or R5-8 can be applied.

\subsubsection{Revisiting the introductory examples}\label{sec:revisit}

We revisit the introductory examples for which we derived testable predictions in section \ref{sec:causterm}.

Firstly, consider the causal hypothesis \textit{amygdala activity} $\to$ \textit{hippocampal activity} $\to$ \textit{explicit memory}.
The two predictions derived in section \ref{sec:causterm}, i.\,e. that \textit{amygdala activity} is marginal dependent on \textit{explicit memory} and becomes conditionally independent given \textit{hippocampal activity}, can be tested by assessing the relevance of \textit{amygdala activity} in a response-based encoding and decoding model, respectively.
Without employing further assumptions these conditions are not sufficient to prove the hypothesized causal structure, though, as \textit{amygdala activity} $\gets$ \textit{hippocampal activity} $\to$ \textit{explicit memory} or \textit{amygdala activity} $\gets$ h $\to$ \textit{hippocampal activity} $\to$ \textit{explicit memory}, where $h$ is hidden, are also consistent with the tested conditions.
A statement that is warranted by interpretation rule R6 in case that both predictions hold true is the following: \textit{amygdala activity} is not a direct cause of \textit{explicit memory}.

Secondly, consider the causal hypothesis \textit{pre-stimulus alpha oscillations} $\to$ \textit{working memory}.
The prediction derived in section \ref{sec:causterm}, i.\,e. \textit{pre-stimulus alpha oscillations} $\centernot\indep$ \textit{working memory}, can be tested by assessing the relevance of \textit{pre-stimulus alpha oscillations} in a response-based encoding model.
Testing this prediction is again not sufficient to conclude the hypothesized causal structure:
in case the prediction holds true, \textit{pre-stimulus alpha oscillations} is only potentially a cause of \textit{working memory} (cf. interpretation rule R1), as \textit{pre-stimulus alpha oscillations} $\gets h \to$ \textit{working memory}, where $h$ is hidden, is also consistent with the tested condition.
Under the rather strong assumption of causal sufficiency the existence of hidden confounders is denied, and the hypothesized causal structure may indeed be concluded.

Note that a hypothesized causal structure can be rejected whenever one of the conditional independences implied by the DAG is not present in the data.

\section{Experimental Results}\label{sec:expres}

In this section, we demonstrate the empirical significance of our theoretical results on EEG data recorded during a visuo-motor learning task. We investigate the neural basis of planning a reaching movement by training encoding and decoding models on EEG bandpower features derived from trial-periods in which subjects have either been instructed to rest or to plan a reaching movement. We chose this stimulus-based setting to illustrate our theoretical results, as it admits less ambiguous causal interpretations than a response-based setting (cf.~Table~\ref{tab:intrule}).

We introduce the experimental setup and recorded data in section \ref{sec:exppar}. For simplicity, we omit details of the experimental setup that are irrelevant for the stimulus-based setting. We then describe the data preprocessing steps and the derivation of our brain state features in section \ref{sec:exppre}. In section \ref{sec:exptrain}, we train a nonlinear encoding and a nonlinear decoding model and determine the set of (ir-)relevant features for each model type. The causal conclusions, that can be derived from our analysis, are presented in section \ref{sec:expind}. We first focus on each model type independently, and then discuss further causal insights that are obtained by combining encoding and decoding models.

\subsection{Experimental setup and data}\label{sec:exppar}
\begin{figure*}
    \centering
        \begin{subfigure}[b]{0.22\textwidth}
            \includegraphics[bb=0 0 617 407,width=\textwidth]{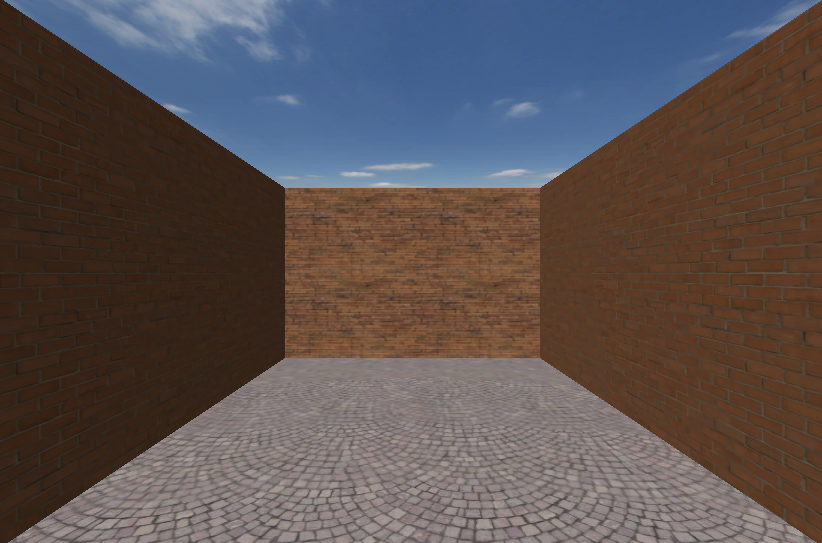}
            \caption{}\label{fig:visPretrial}
        \end{subfigure}%
        \vspace{0.03\textwidth}
        \begin{subfigure}[b]{0.22\textwidth}
            \includegraphics[bb=0 0 621 409,width=\textwidth]{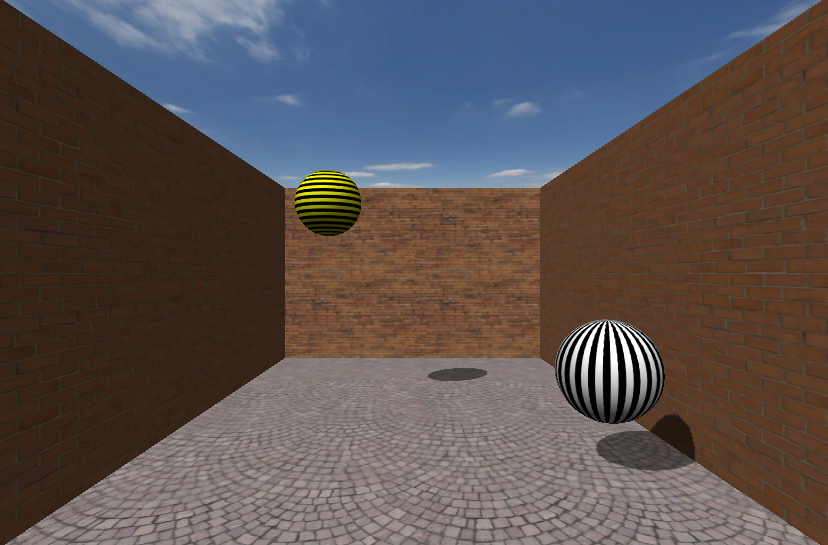}
            \caption{}\label{fig:visPlanning}
        \end{subfigure}%
        \vspace{0.03\textwidth}
        \begin{subfigure}[b]{0.22\textwidth}
            \includegraphics[bb=0 0 620 413,width=\textwidth]{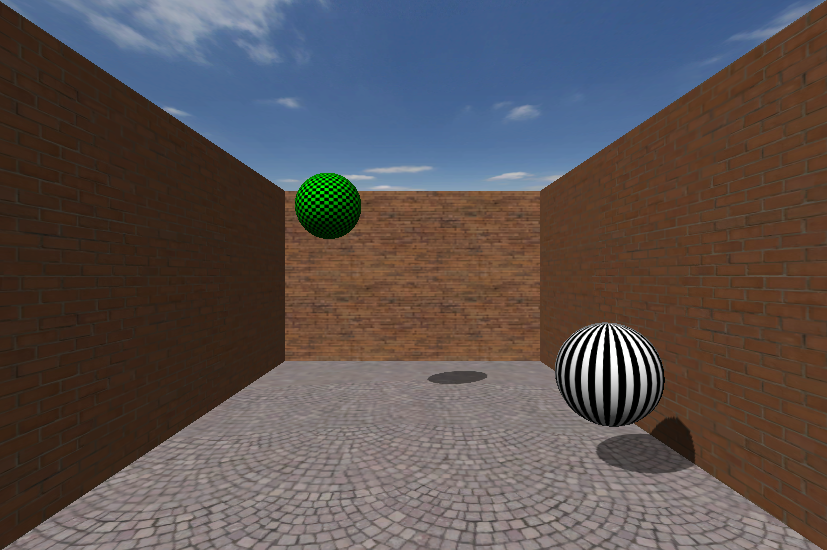}
            \caption{}\label{fig:visReaching}
        \end{subfigure}%
        \vspace{0.03\textwidth}
        \begin{subfigure}[b]{0.22\textwidth}
            \includegraphics[bb=0 0 621 411,width=\textwidth]{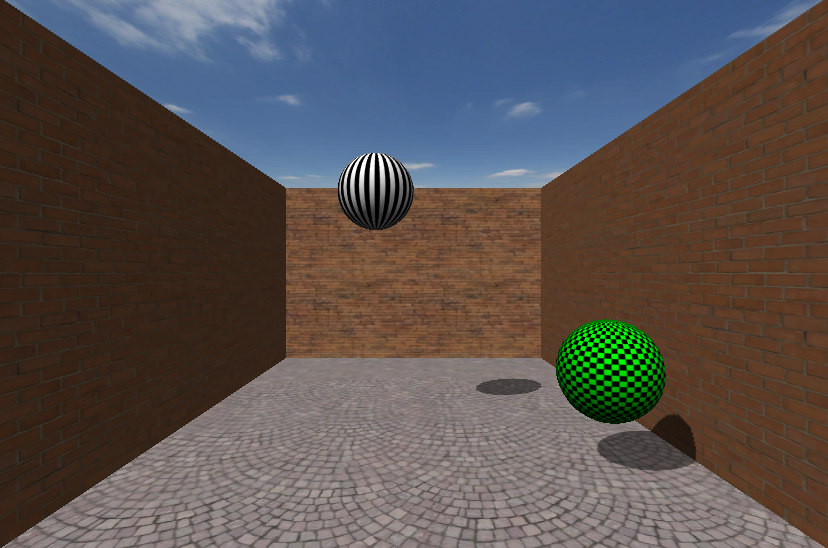}
            \caption{}\label{fig:visReturn}
        \end{subfigure}\\
    \caption{Visual feedback used during different phases of a trial.
            (a) Pre-trial phase: the blank 3D virtual reality scene instructs the subject to relax.
            (b) Planning phase: white and yellow balls appear, representing the position of the subject's right arm and the target location, respectively, instructing the subject to plan but not yet initiate a reaching movement to the indicated location.
            (c) Reaching phase: the target's color switches to green, instructing the subject to perform the reaching movement.
            (d) Return phase: having completed the reaching movement, the subject is instructed to return to the initial starting position, indicated by the green ball.}\label{fig:vis}
\end{figure*}

Subjects were trained in a visuo-motor learning paradigm to maximize the smoothness of three-dimensional reaching movements performed with their right hand. Specifically, subjects were placed approximately 1.5 m in front of a computer screen displaying a 3D virtual reality scene. At the beginning of each trial, subjects were shown the blank 3D virtual reality scene, instructing them to rest (figure \ref{fig:visPretrial}). After five seconds, subjects were shown the position of their arm, indicated by a ball with a black and white checkerboard texture, and the target for the upcoming reaching movement, indicated by a ball with a yellow and black checkerboard texture (figure \ref{fig:visPlanning}). This instructed subjects to plan but not yet initiate the reaching movement. After a variable delay of 2.5 to 4 seconds, drawn from a uniform distribution, the yellow ball turned green, instructing subjects to carry out the movement (figure \ref{fig:visReaching}). At the end of each trial, subjects received feedback on the smoothness of their reaching movement, derived from position measurements recorded by a motion tracking device (Impulse X2 Motion Capture System, PhaseSpace, San Leandro, CA, U.S.), and were instructed to return to the initial starting position (figure \ref{fig:visReturn}). Each subject carried out a total of 250 movements to randomly chosen locations within the subject's movement range, distributed across 5 blocks with brief intermissions in between.

Throughout the training session, a 120-channel EEG, with electrodes placed according to the extended 10-20 system, was recorded at 500 Hz sampling rate using actiCAP electrodes and four BrainAmp DC amplifiers (BrainProducts, Gilching, Germany). Prior to each training session, a five-minute resting-state baseline was recorded, for which subjects were instructed to relax while looking at a fixation cross displayed centrally on the screen. All recordings were carried out with a reference electrode behind the left ear and converted to common average reference offline.

Eighteen healthy male subjects (mean age $28.32\pm 8.48$) participated in this study. All subjects indicated that they are right-handed and conducted the study with their right arm.
Subjects gave informed consent in accordance with guidelines set by the Max Planck Society. The study was approved by an ethics committee of the Max Planck Society.

\subsection{Data preprocessing and feature computation}\label{sec:exppre}

In order to facilitate an across-subject analysis, we chose to train our models on bandpower features derived from a group-wise independent component analysis (ICA).
This is particularly helpful as it allows to reduce the dimensionality of the feature space and leads to physiologically interesting sources. We note that this cannot introduce artificial dependences between subjects' trial data as long as the ICA spatial filters obtained from resting state data are being held fixed throughout further analysis. Furthermore, although the group-wise ICA seeks independent components in the resting state data, the dependency structure between those components may be very different during a task.

\begin{figure*}
        \centering
        \begin{subfigure}[b]{0.3\textwidth}
            \includegraphics[width=\textwidth,bb=0 0 448 336]{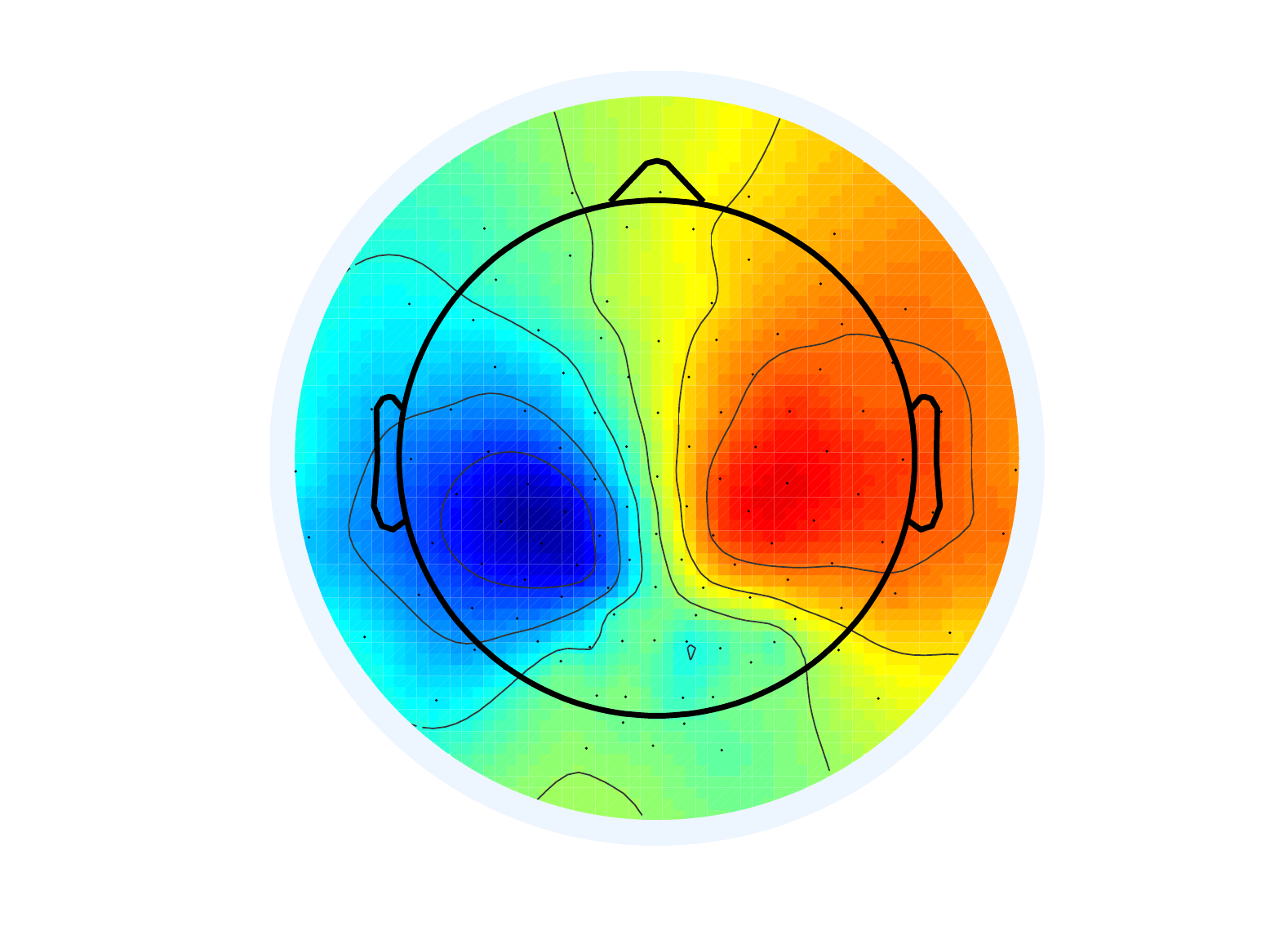}%
        \end{subfigure}%
        \vspace{0.03\textwidth}
        \begin{subfigure}[b]{0.3\textwidth}
            \includegraphics[width=\textwidth,bb=0 0 448 337]{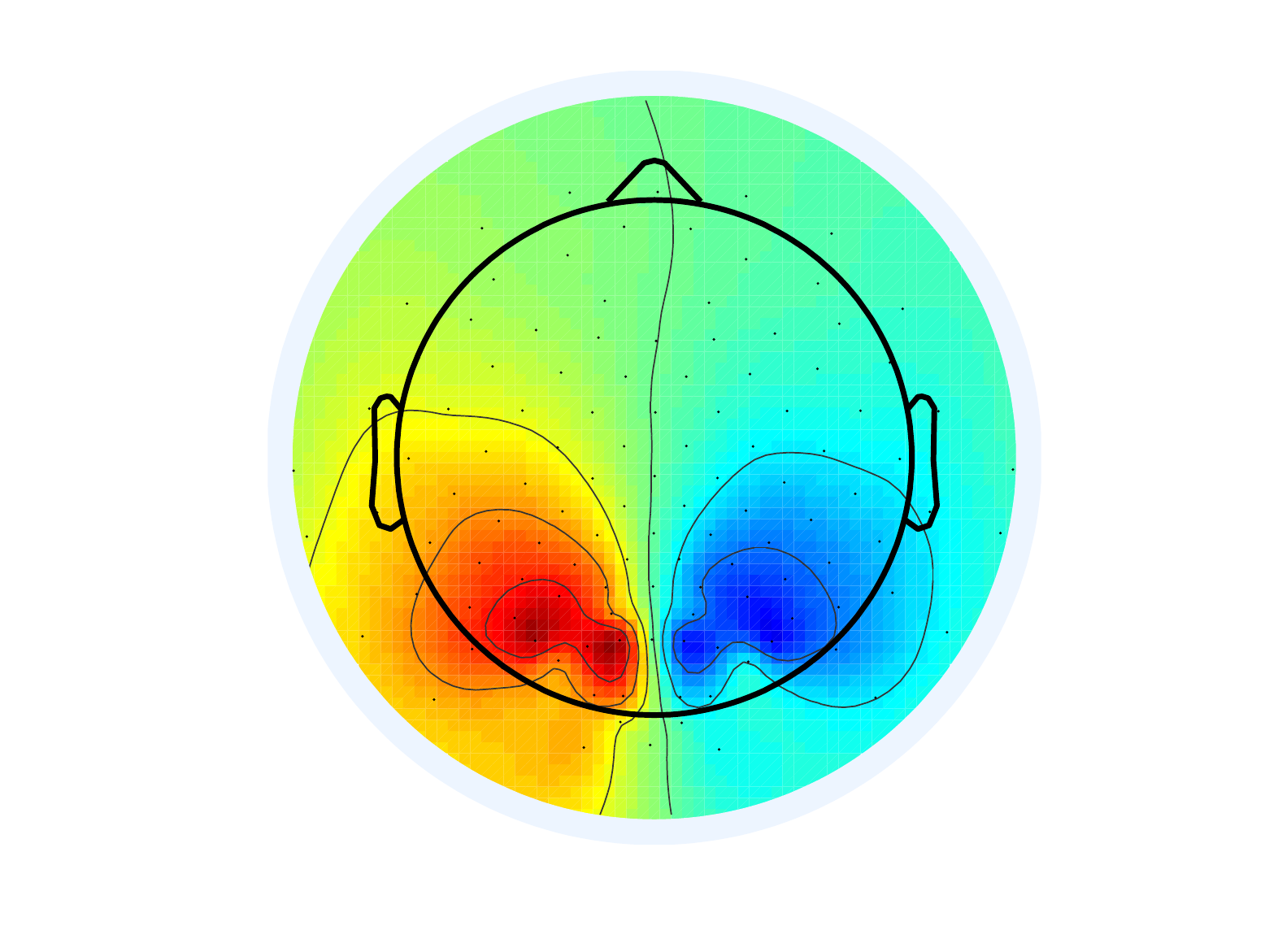}
        \end{subfigure}%
        \vspace{0.03\textwidth}
        \begin{subfigure}[b]{0.3\textwidth}
            \includegraphics[width=\textwidth,bb=0 0 448 337]{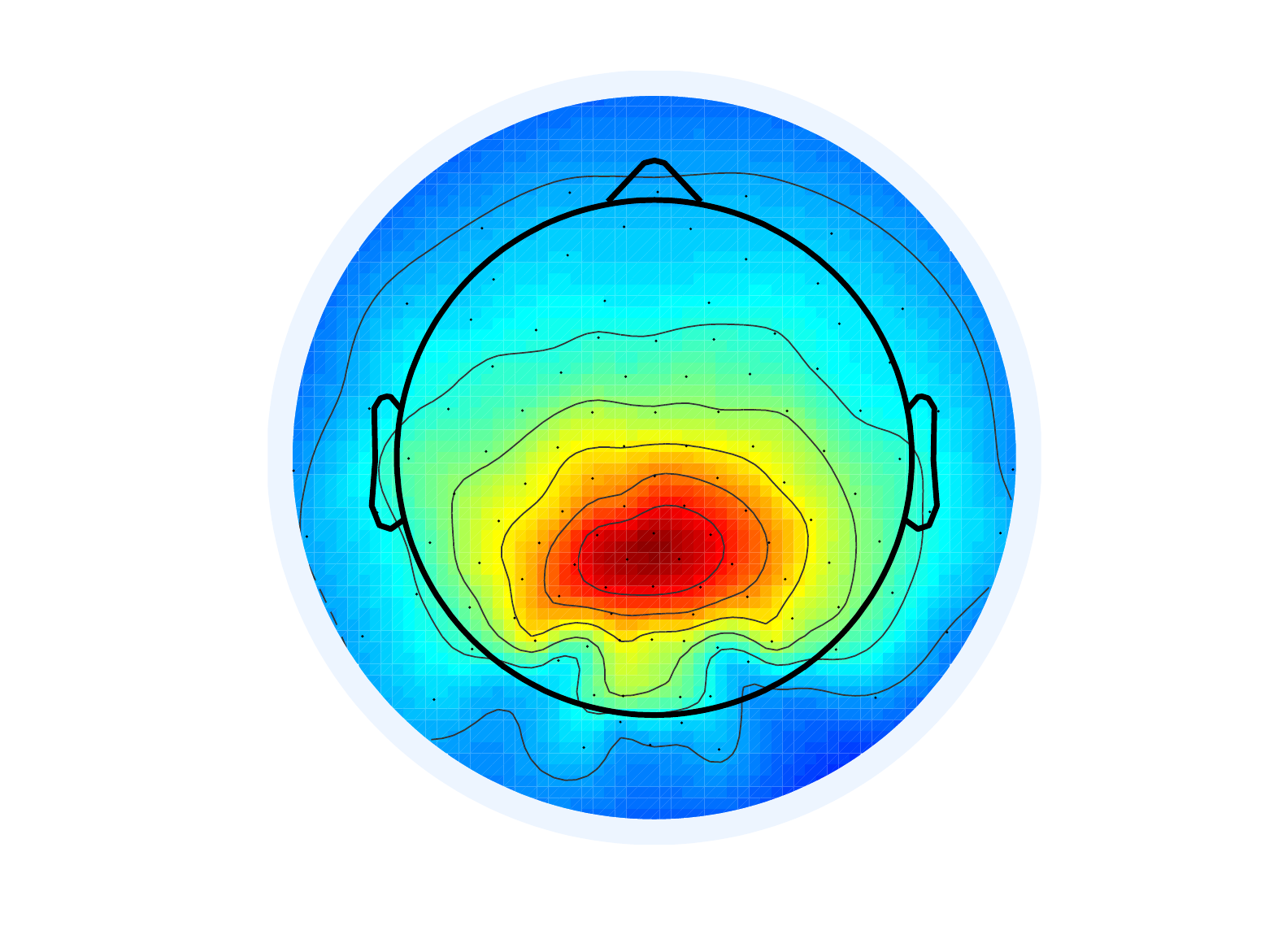}
        \end{subfigure}\\

        \begin{subfigure}[b]{0.3\textwidth}
            \includegraphics[bb=0 0 1058 872,width=\textwidth]{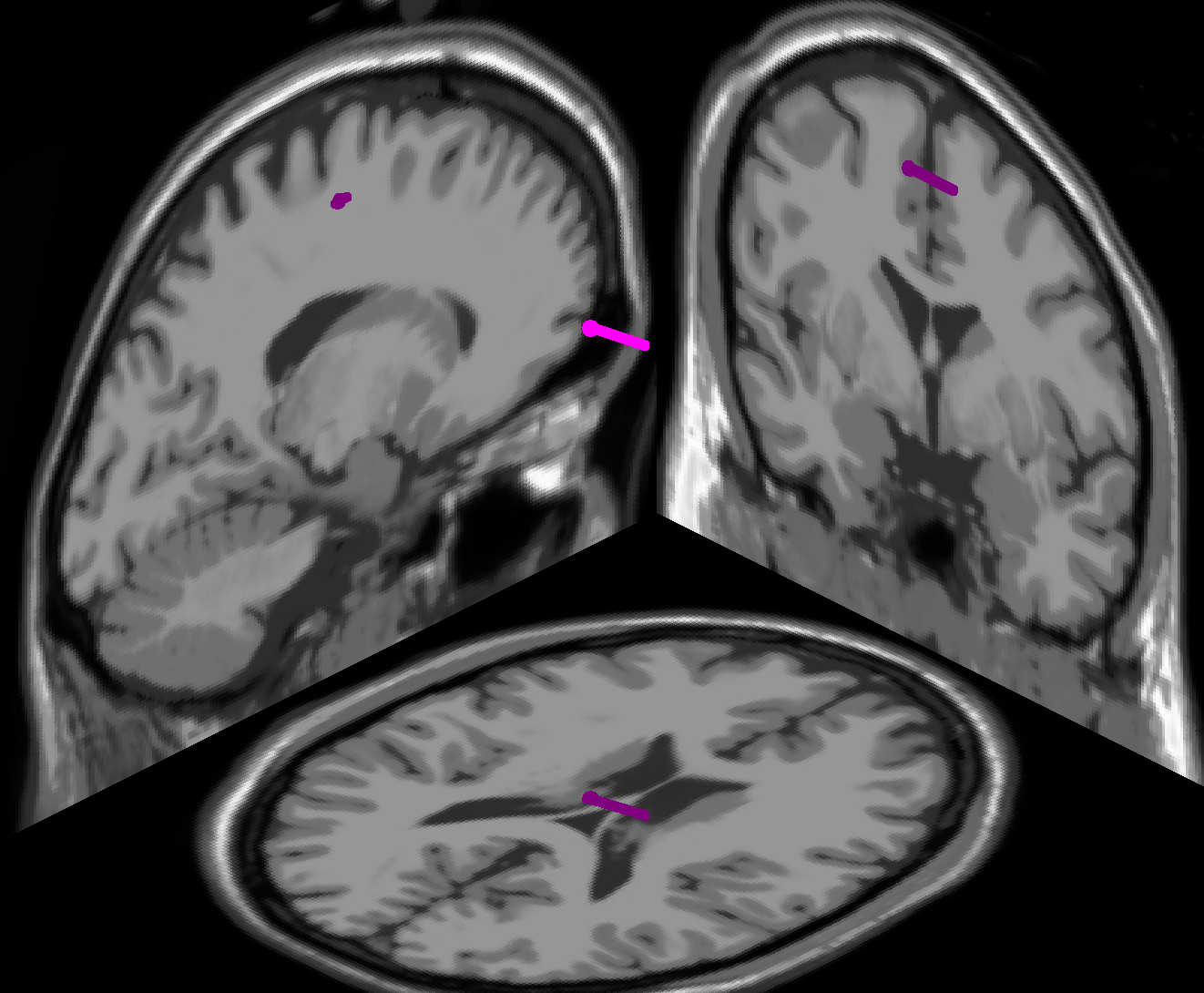}%
            \caption{IC 1, RV 5.88\%}
            \label{fig:IC1}
        \end{subfigure}%
        \vspace{0.03\textwidth}
        \begin{subfigure}[b]{0.3\textwidth}
            \includegraphics[bb=0 0 1058 872,width=\textwidth]{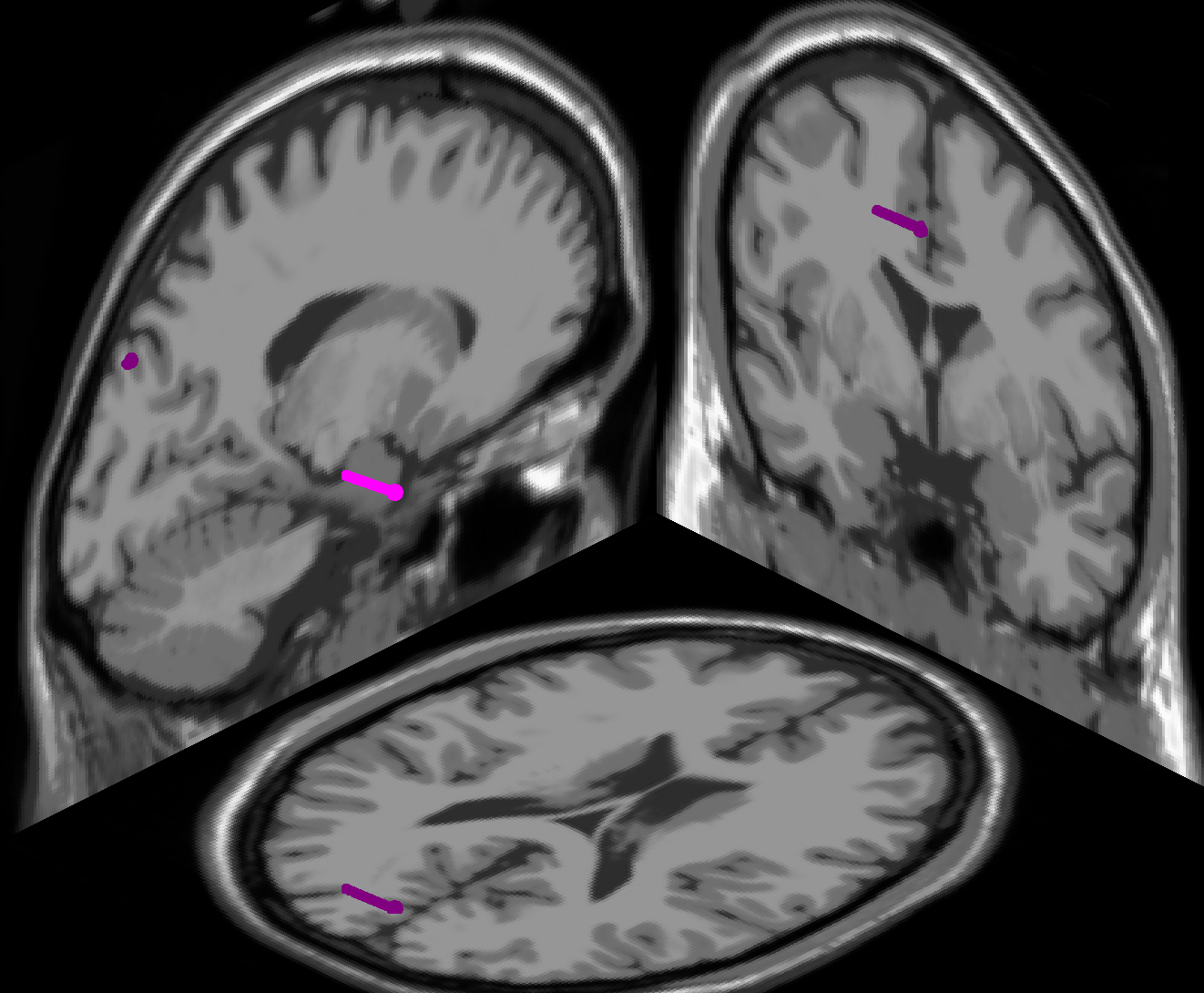}
            \caption{IC 2, RV 3.28\%}
            \label{fig:IC2}
        \end{subfigure}%
        \vspace{0.03\textwidth}
        \begin{subfigure}[b]{0.3\textwidth}
            \includegraphics[bb=0 0 1058 872,width=\textwidth]{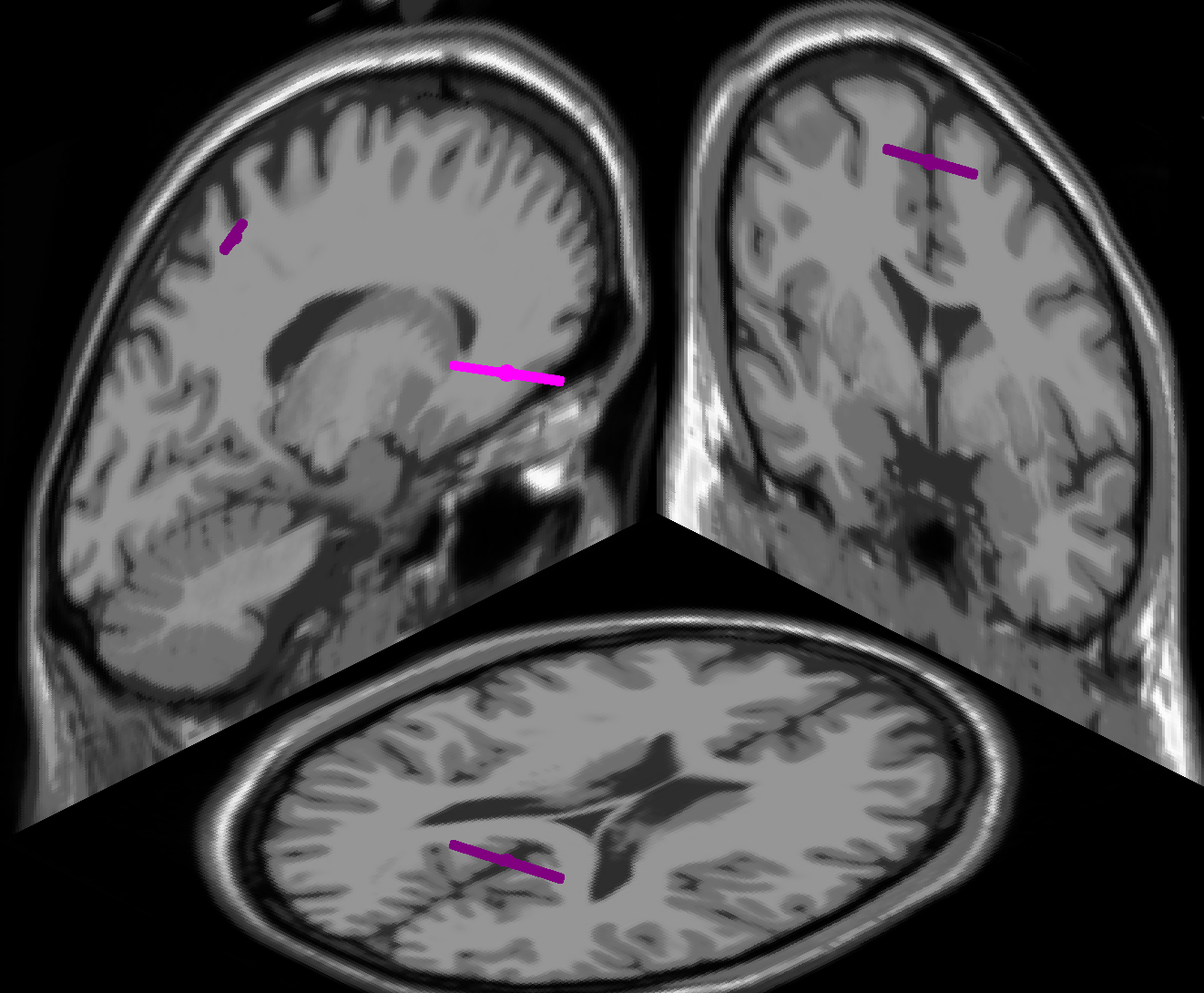}
            \caption{IC 3, RV 4.39\%}
            \label{fig:IC3}
        \end{subfigure}\\

        \begin{subfigure}[b]{0.3\textwidth}
            \includegraphics[bb=0 0 448 337,width=\textwidth]{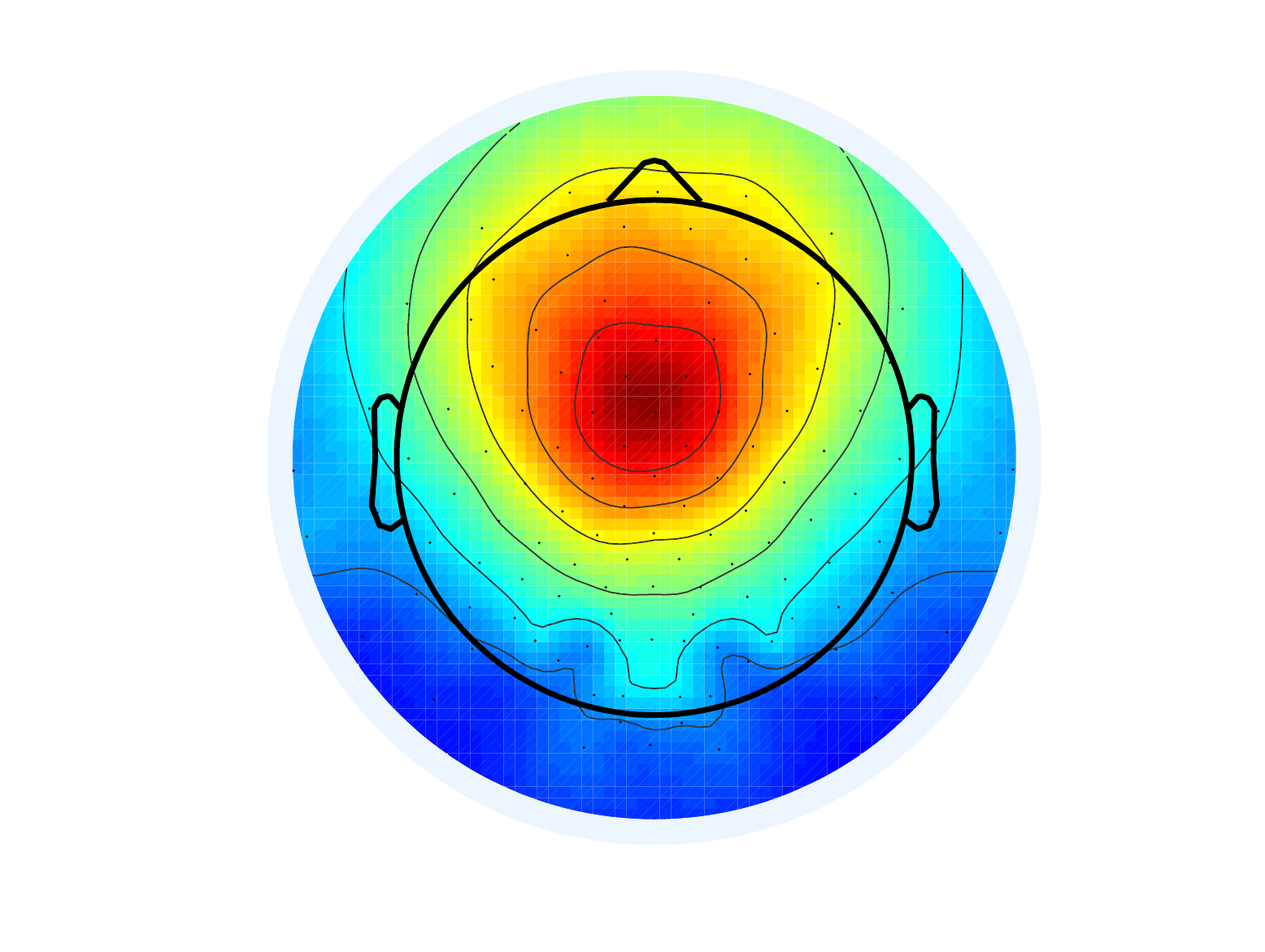}
        \end{subfigure}%
        \vspace{0.03\textwidth}
        \begin{subfigure}[b]{0.3\textwidth}
            \includegraphics[bb=0 0 448 337,width=\textwidth]{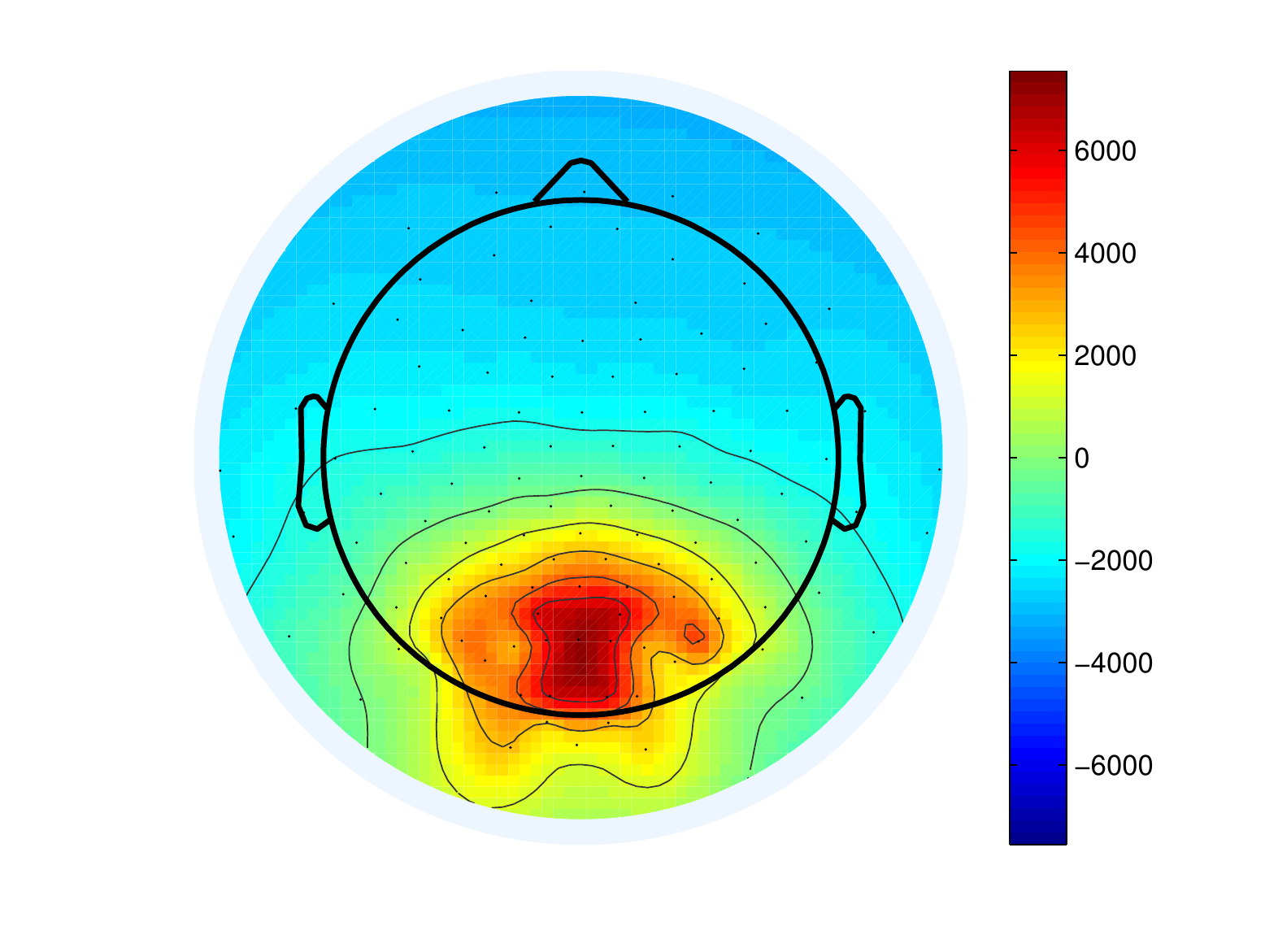}%
        \end{subfigure}%
        \vspace{0.03\textwidth}
        \begin{subfigure}[b]{0.3\textwidth}
            \includegraphics[bb=0 0 448 337,width=\textwidth]{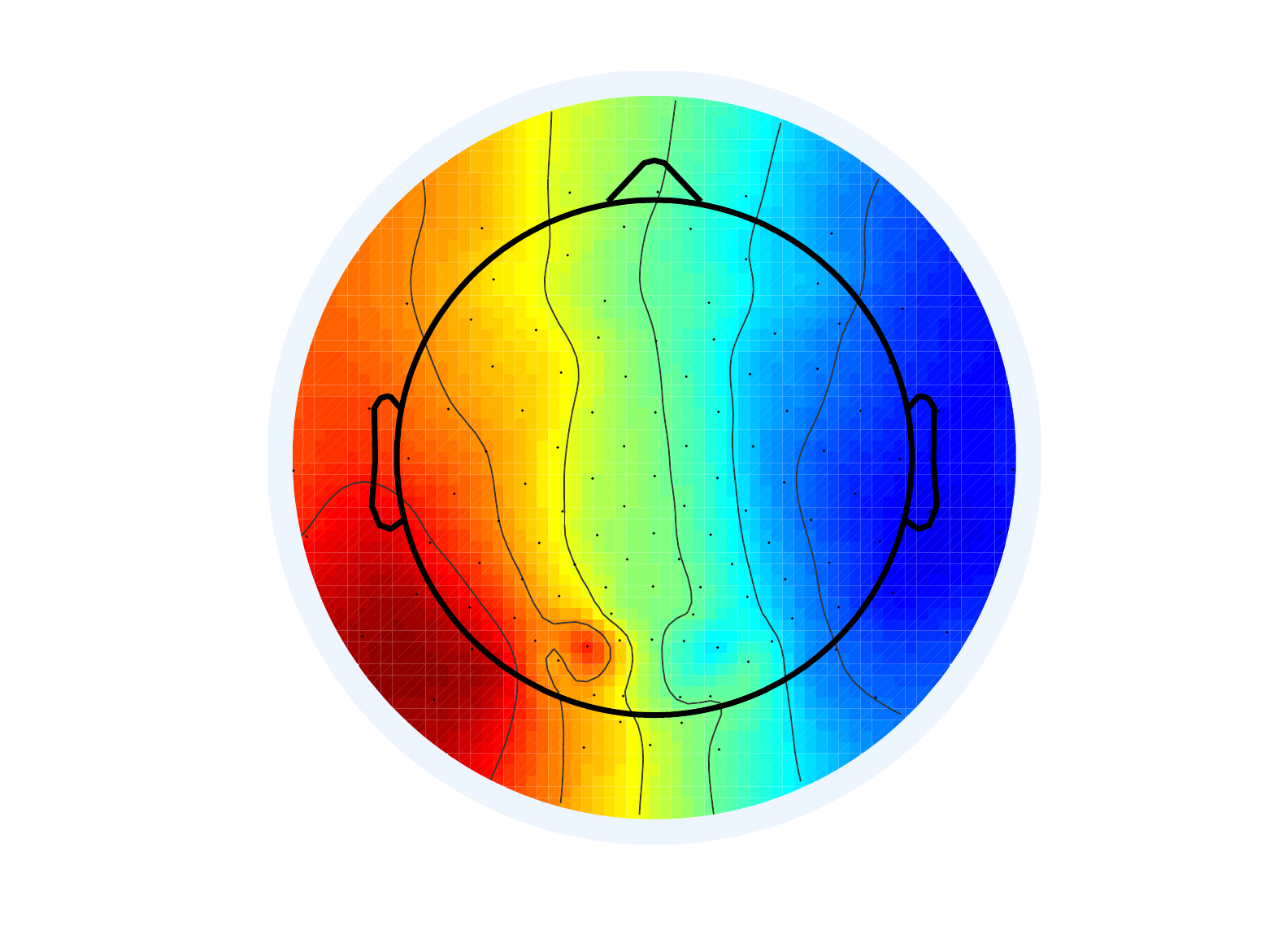}%
        \end{subfigure}\\

        \begin{subfigure}[b]{0.3\textwidth}
            \includegraphics[bb=0 0 1058 872,width=\textwidth]{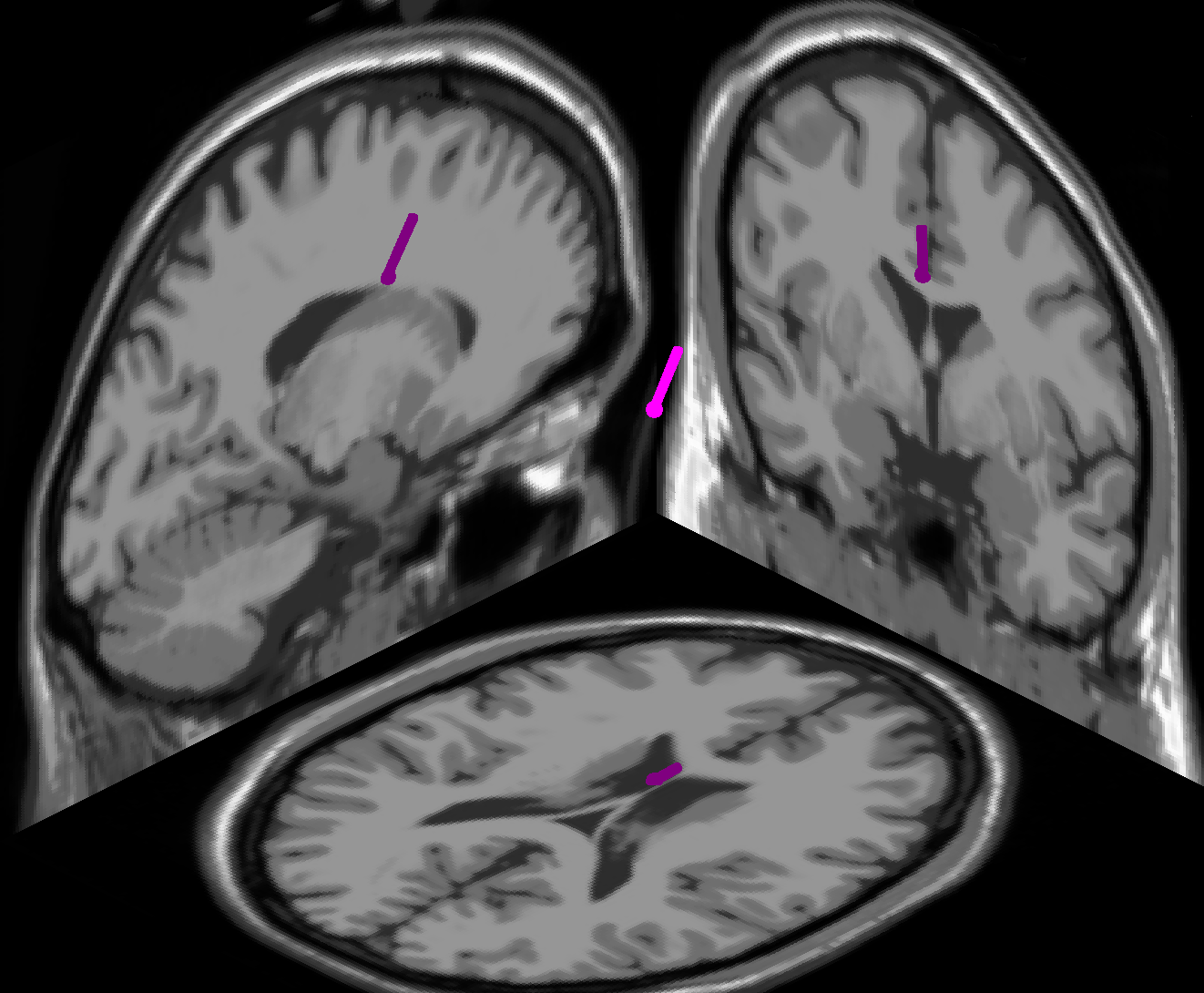}
            \caption{IC 4, RV 3.25\%}
            \label{fig:IC4}
        \end{subfigure}%
        \vspace{0.03\textwidth}
        \begin{subfigure}[b]{0.3\textwidth}
            \includegraphics[bb=0 0 1058 872,width=\textwidth]{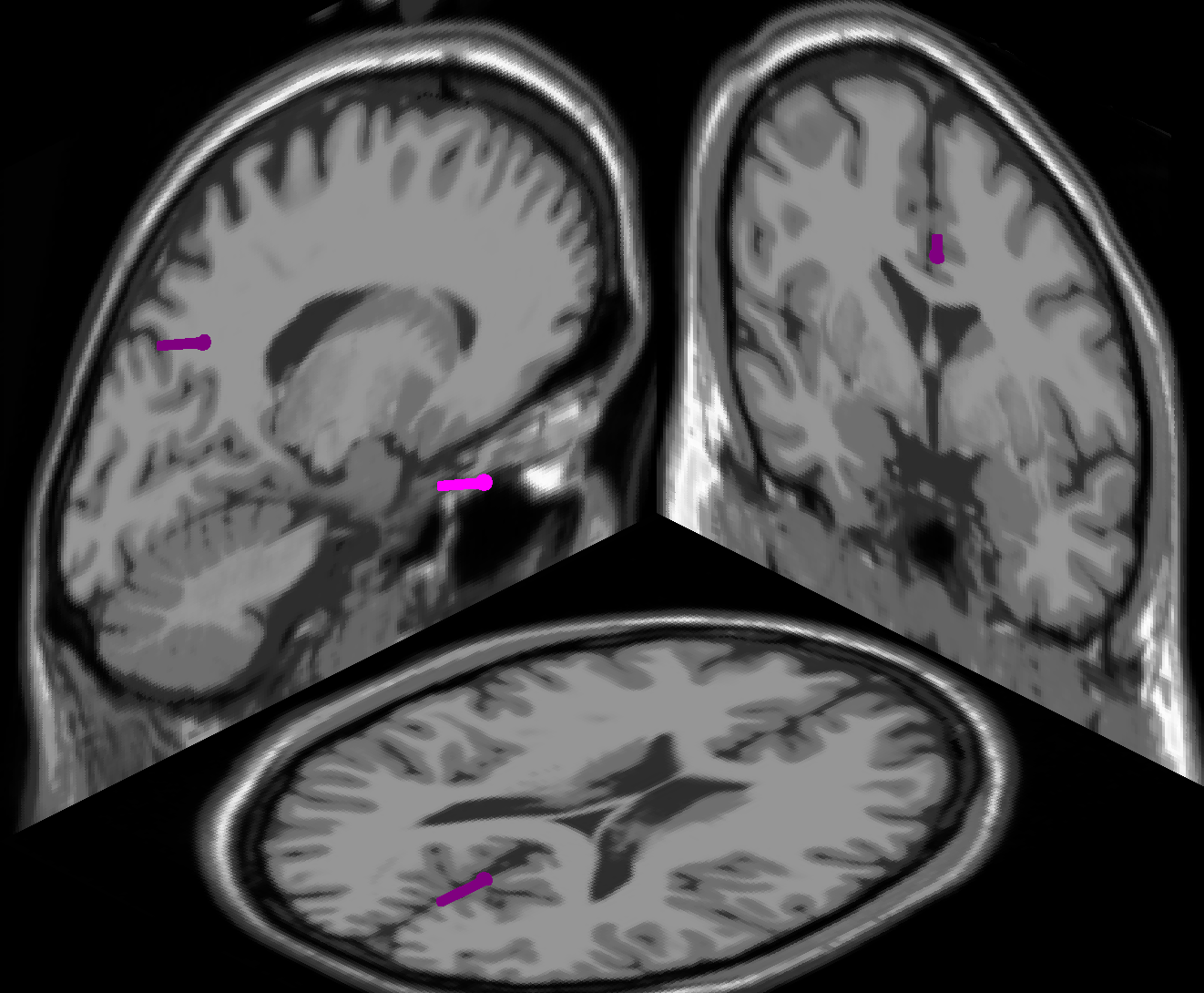}%
            \caption{IC 5, RV 3.07\%}
            \label{fig:IC5}
        \end{subfigure}%
        \vspace{0.03\textwidth}
        \begin{subfigure}[b]{0.3\textwidth}
            \includegraphics[bb=0 0 1058 872,width=\textwidth]{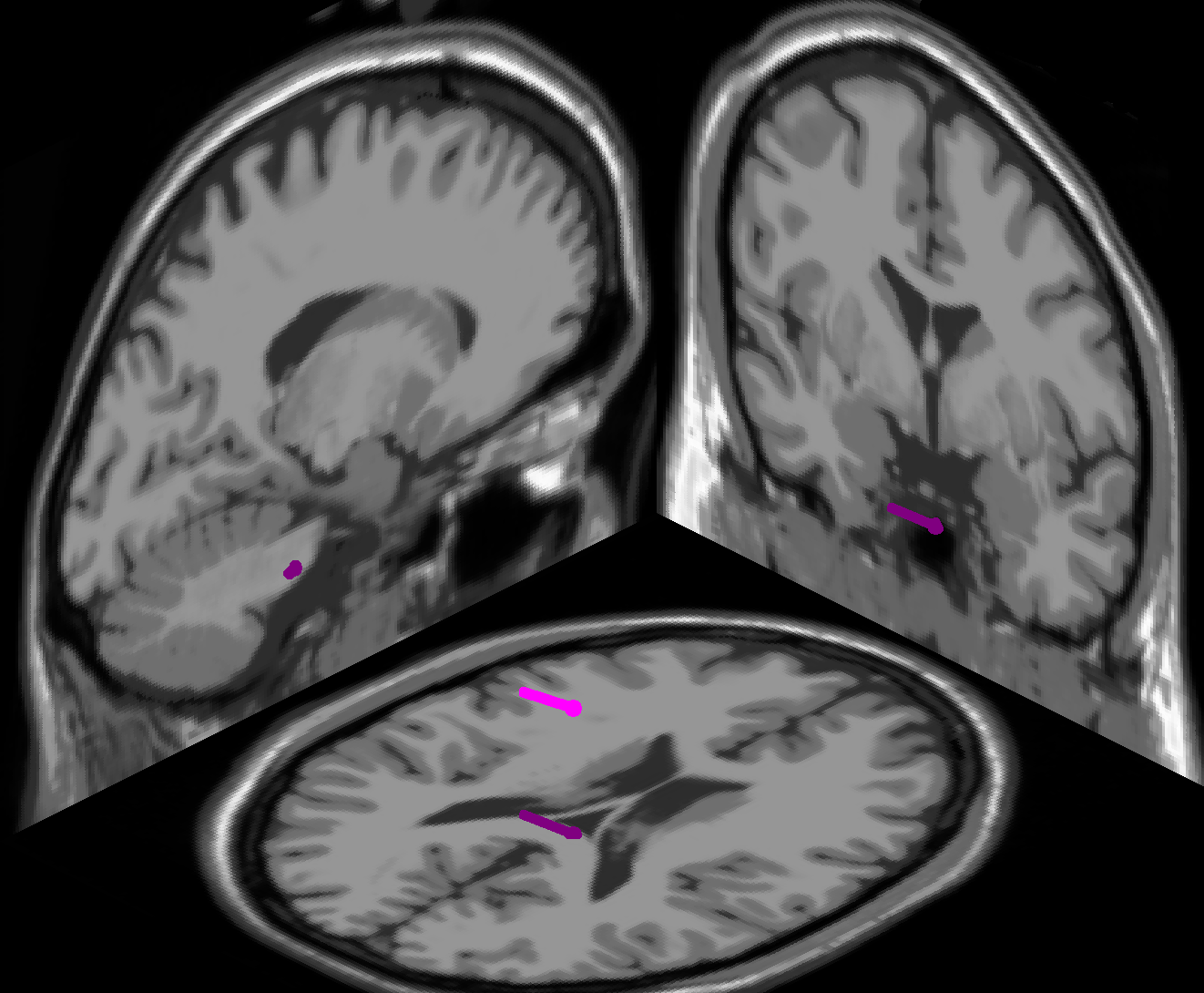}%
            \caption{IC 6, RV 1.88\%}
            \label{fig:IC6}
        \end{subfigure}\\

        \caption{Topographies and equivalent dipole locations of the six cortical ICs from the group-wise ICA. The residual variances of the equivalent dipole fits are denoted by RV.}\label{fig:ics}
\end{figure*}

We first high-pass filtered the resting-state data of each subject at a cut-off frequency of 3 Hz, and then performed a group-wise ICA on the combined resting-state data of all subjects.
We chose the SOBI algorithm \cite{Belouchrani:1997} for this purpose, as ICA algorithms based on second-order statistics have been shown to outperform methods based on higher moments in group-wise analyses \cite{Lio:2013}. We manually inspected the topography and spectrum of every resulting independent component (IC) and rejected all ICs as non-cortical that did not exhibit a clear dipolar topography \cite{Delorme:2012}.
The topographies and equivalent dipoles of the six cortical ICs that we retained for further analyses are shown in figure \ref{fig:ics}.
Equivalent dipole locations were derived with a three-shell spherical head model with standardized electrode locations, using the EEGLAB toolbox \cite{Delorme:2004}. We note that ICs 1 and 2, located in sensorimotor and occipital cortex, represent sensorimotor and low-level visual processes, respectively. In contrast, ICs 3, 4 and 5, located in precuneus, in the anterior cingulate, and at the intersection of cuneus and precuneus, respectively, are generated in cortical areas linked to fronto-parietal attention networks \cite{Bressler:2010}. IC 6 appears to represent a sub-cortical source.

To compute the brain state features for the causal analysis, we first applied the spatial filters of the six ICs in figure \ref{fig:ics} to each subject's EEG data recorded during the training session, and then computed log-bandpower in the $\alpha$-range (8--14 Hz) for each subject, IC, and trial in two temporal windows. For the first window, we selected a random 1.5 s window from each trial's rest phase, excluding the first and last 500 ms to avoid carry-over effects. For the second window, we selected a random 1.5 s window from each trial's planning phase, again avoiding the first and last 500 ms. Trials, in which subjects already initiated the movement during the planning phase or did not reach the target within 10 s, were excluded as invalid trials. We limited our analysis to the $\alpha$-band, as we found $\alpha$-rhythms most relevant for predicting the progress of visuo-motor learning in a previous study \cite{Meyer:2014}. We excluded one subject from further analyses, as we found this subject's spectra to be contaminated by EMG activity.

For each of the remaining 17 subjects, we thus obtained between 444 and 498 samples of a six-dimensional feature vector, representing log-bandpower in the $\alpha$-range of six cortical ICs in windows of 1.5 s length. In half of these samples, subjects had been instructed to rest, while in the other half subjects had been instructed to plan a reaching movement.

\subsection{Model training and feature analysis}\label{sec:exptrain}

In the present context, the experimental condition $C$ corresponds to the instruction to either rest or to plan a reaching movement. As this instruction always preceded the time-window used to compute the corresponding brain state features, we are working in a stimulus-based setting, i.\,e.~$C \equiv S$ (cf.~section~\ref{sec:causalmodels}). Our brain state features $X = \{X_1,...,X_6\}$ correspond to log-bandpower in the $\alpha$-range of the cortical ICs in figure \ref{fig:ics}. In the following, we first describe how we trained a nonlinear encoding and a nonlinear decoding model to distinguish between relevant and irrelevant brain state features in each model type. The models were trained on each subject's features individually. We then discuss how we combined the experimental results across subjects to identify the \mbox{(ir-)relevant} feature sets on a group-level. These sets form the basis for the causal interpretation in section~\ref{sec:expind}.

\subsubsection{Encoding analysis}
To quantify the relevance of features in encoding, we employed a non-linear independence test based on the Hilbert–Schmidt independence criterion (HSIC) \cite{gretton2008kernel,Grosse-WentrupNeuroImage2011}. The HSIC test utilizes a kernel independence measure, capable of detecting linear as well as non-linear dependences between arbitrary input variables. The p-values of the HSIC tests were computed using a permutation analysis with $1000$ permutations. The kernel size for the HSIC tests was set to the median distance between points in input space \cite{gretton2008kernel}. For each subject and feature, this resulted in a p-value that quantifies the probability that the observed data has been generated under the precondition $S \indep X_i$ (cf.~Table~\ref{tab:encres}).

\begin{table*}[t]
    \small
    \centering

    \caption{p-values that quantify the probability of $S \indep X_i$ for each subject and feature. Values exceeding $0.05$ are highlighted in gray. The p-values of the Kolmogorov-Smirnov permutation test, that quantify the probability of the values of each IC being drawn from a uniform distribution, are denoted by \emph{KSp}.}
    \centering
    \tabcolsep=0.15cm
\begin{tabular}{lllllll}
    Subject & IC 1 & IC 2 & IC 3 & IC 4 & IC 5 & IC 6 \\
\hline
1 &0 &0 &0 &0 &0 &0\\
2 &0 &0 &0 &0.017 &0 &0\\
3 &0 &0 &0 &0.006 &0&\cellcolor{gray!25}0.522\\
4 &0 &0.029&\cellcolor{gray!25}0.151 &0.002&\cellcolor{gray!25}0.192 &0.002\\
5 &0 &0 &0 &0.012 &0.003 &0.037\\
6 &0 &0 &0 &0.01 &0&\cellcolor{gray!25}0.202\\
7 &0 &0 &0 &0 &0 &0\\
8 &0 &0 &0 &0&\cellcolor{gray!25}0.061 &0\\
9&\cellcolor{gray!25}0.061 &0&\cellcolor{gray!25}0.097&\cellcolor{gray!25}0.265 &0.012&\cellcolor{gray!25}0.596\\
10 &0 &0 &0 &0.008 &0 &0.001\\
11 &0 &0 &0 &0.005 &0 &0.036\\
12 &0 &0 &0.001&\cellcolor{gray!25}0.07 &0.001 &0\\
13 &0 &0 &0 &0.013 &0&\cellcolor{gray!25}0.217\\
14 &0 &0.027&\cellcolor{gray!25}0.055&\cellcolor{gray!25}0.363&\cellcolor{gray!25}0.102&\cellcolor{gray!25}0.246\\
15 &0 &0 &0 &0.025 &0.005 &0\\
16 &0 &0 &0 &0 &0 &0.032\\
17&\cellcolor{gray!25}0.357 &0.024 &0.019&\cellcolor{gray!25}0.749 &0.003 &0.001\\
\hline
KSp & 0 & 0 & 0 & 0 & 0 & 0\\
\end{tabular}
    \label{tab:encres}
\end{table*}

\subsubsection{Decoding analysis}
We used a random forest classifier (RF) \cite{breiman2001random} to decode the experimental condition from the brain state features. For each RF model, $100$ trees were grown using bagging, i.\,e.~for every tree a training subset ($63.2\%$ of the whole training set) was drawn by sampling with replacement. While growing the trees, at each node the best split of $\lfloor\sqrt{6}\rfloor = 2$ (as suggested in \cite{breiman2002manual}) randomly chosen features was used. Estimates of prediction accuracy (PE*) were obtained by leave-one-trial-out cross-validation. This resulted in the subject-specific decoding accuracies shown in Table \ref{tab:pestar}.
A Wilcoxon signed rank test rejected the null-hypothesis of chance-level decoding accuracies on the group-level with $p = 2.9305e-04$.

\begin{table}[h]
    \small
    \centering

    \caption{Prediction accuracy estimates for every subject in percentage of correctly decoded trials with all features intact (PE*).}
    \centering
    \tabcolsep=0.15cm
\begin{tabular}{l|rrrrrrrrrrrrrrrrr}
    Subject & 1 & 2 & 3 & 4 & 5 & 6 & 7 & 8 & 9 \\ \hline
    PE* & 83.83 & 69.92 & 73.08 & 60.14 & 60.37 & 71.11 & 74.15 & 73.39 & 58.61 \\
    \hline \hline
    Subject & 10 & 11 & 12 & 13 & 14 & 15 & 16 & 17 &\\
    PE* & 72.28 & 82.34 & 65.79 & 74.19 & 60.82 & 66.03 & 67.95 & 53.11 &\\
\end{tabular}
    \label{tab:pestar}
\end{table}

To quantify the relevance of features in decoding, we then successively permuted every feature $1000$ times and compared the resulting PEs to the PE with all features intact (PE*).  For each subject and feature, this resulted in a p-value that quantifies the probability that the observed data has been generated under the precondition $S \indep X_i | X \setminus X_i$ (cf.~Table~\ref{tab:decres}).

\subsubsection{Identifying (ir-)relevant feature sets on the group-level}
In order to identify the (ir-)relevant features on the group-level, we make use of the fact that by construction p-values are uniformly distributed under the null-hypothesis.
In the present context, this implies that if a feature is irrelevant in encoding/decoding, the 17 subject-specific p-values of this feature are drawn from a uniform distribution.
This enables us to use a Kolmogorov-Smirnov permutation test (KS test) (with $10^5$ permutations), for each feature and model type, to quantify the probability of the set of observed p-values under the assumption of (conditional) independence. We reject the null-hypothesis of $S \indep X_i$ or $S \indep X_i | X \setminus X_i$, for the encoding or decoding side, respectively, if the resulting p-value of the KS test is below the significance level $\alpha = 0.05$. Conversely, we accept the null-hypothesis if the p-value across subjects exceeds $\beta = 0.10$. In this way, we can test each feature's significance for encoding and decoding on a group-level, while avoiding problems in causal analysis related to pooling features across multiple subjects \cite{Ramsey:2010}.

The resulting p-values are shown in Table \ref{tab:encres} and \ref{tab:decres} for the encoding and decoding analysis, respectively. For the encoding side, we find all six cortical ICs to be relevant. For the decoding side, we only find ICs 1 and 2 to be relevant. We hence obtain the following sets of (ir-)relevant features on the group-level:
\begin{itemize}
    \item $X^\text{+enc} = \{X_1,...,X_6\}$
    \item $X^\text{--enc} = \emptyset$
    \item $X_\text{+dec} = \{X_1,X_2\}$
    \item $X_\text{--dec} = \{X_3,X_4,X_5,X_6\}$
\end{itemize}
This enables us to distinguish between ICs that are relevant only for encoding and ICs that are relevant in both model types: $X^\text{+enc}_\text{--dec} = \{X_3,X_4,X_5,X_6\}$ and $X^\text{+enc}_\text{+dec} = \{X_1,X_2\}$.

\subsection{Causal analysis}\label{sec:expind}
Having identified the sets of (ir-)relevant features, we proceed to apply our causal interpretation rules.
We first consider each model type independently, and then discuss causal conclusions that can be derived from considering both models jointly.

\subsubsection{Encoding model only}
From $X^\text{+enc} = \{X_1,...,X_6\}$ and interpretation rule S1, it follows that $S$ is causal for the $\alpha$-bandpower of every IC. Accordingly, we find that every IC in figure \ref{fig:ics} responds to the instruction to plan a reaching movement. The specific role of each of these cortical processes, however, remains speculative.

\subsubsection{Decoding model only}
The above chance-level decoding accuracy on the group-level indicates that at least one of the six ICs is an effect $S$. Interpretation rules S3 and S4, however, forbid any further causal interpretation.

\subsubsection{Combination of the encoding and the decoding model}
From $X^\text{+enc}_\text{--dec} = \{X_3,X_4,X_5,X_6\}$ in combination with interpretation rule S6, we conclude that ICs 3--6 are only indirect effects of $S$.
As $X^\text{+enc}_\text{+dec} = \{X_1,X_2\}$, we can further conclude that at least one of ICs 1 and 2 is a direct effect of $S$. This follows from the fact that if $X_1$ and $X_2$ were both indirect effects, all observed brain state features would be indirect effects.

We note that these causal conclusions are in line with the roles commonly attributed to the corresponding cortical areas. Specifically, the higher-level cortical areas represented by ICs 3--5 are only found to be indirectly modulated by the instruction to plan a reaching movement. In contrast, low-level sensorimotor and visual areas, represented by ICs 1 and 2, respectively, are found to contain at least one direct effect of the instruction to plan a movement.

\begin{table*}[t]
    \small
    \centering

    \caption{p-values that quantify the probability of $S \indep X_i | X \setminus X_i$ for each subject and feature. Values exceeding $0.05$ are highlighted in gray. The p-values of the Kolmogorov-Smirnov permutation test, that quantify the probability of the values of each IC being drawn from a uniform distribution, are denoted by \emph{KSp}.}
    \centering
    \tabcolsep=0.15cm
\begin{tabular}{lllllll}
    Subject & IC 1 & IC 2 & IC 3 & IC 4 & IC 5 & IC 6 \\
\hline
1&\cellcolor{gray!25}0.82&\cellcolor{gray!25}0.167&\cellcolor{gray!25}0.804&\cellcolor{gray!25}0.37 &0.023&\cellcolor{gray!25}0.261\\
2 &0.022 &0&\cellcolor{gray!25}0.589&\cellcolor{gray!25}0.812&\cellcolor{gray!25}0.655&\cellcolor{gray!25}0.573\\
3 &0.002 &0&\cellcolor{gray!25}0.632&\cellcolor{gray!25}0.233&\cellcolor{gray!25}0.85&\cellcolor{gray!25}0.354\\
4 &0.036&\cellcolor{gray!25}0.159&\cellcolor{gray!25}0.58&\cellcolor{gray!25}0.161&\cellcolor{gray!25}0.392&\cellcolor{gray!25}0.511\\
5&\cellcolor{gray!25}0.658 &0&\cellcolor{gray!25}0.505&\cellcolor{gray!25}0.401&\cellcolor{gray!25}0.645&\cellcolor{gray!25}0.646\\
6&\cellcolor{gray!25}0.08 &0&\cellcolor{gray!25}0.419&\cellcolor{gray!25}0.512&\cellcolor{gray!25}0.794&\cellcolor{gray!25}0.694\\
7&\cellcolor{gray!25}0.259 &0 &0.014&\cellcolor{gray!25}0.414&\cellcolor{gray!25}0.139&\cellcolor{gray!25}0.529\\
8 &0 &0&\cellcolor{gray!25}0.551&\cellcolor{gray!25}0.506&\cellcolor{gray!25}0.542 &0.046\\
9&\cellcolor{gray!25}0.802 &0.034&\cellcolor{gray!25}0.134&\cellcolor{gray!25}0.336&\cellcolor{gray!25}0.691&\cellcolor{gray!25}0.725\\
10&\cellcolor{gray!25}0.233 &0&\cellcolor{gray!25}0.805&\cellcolor{gray!25}0.723&\cellcolor{gray!25}0.406&\cellcolor{gray!25}0.611\\
11 &0&\cellcolor{gray!25}0.122 &0.043&\cellcolor{gray!25}0.406&\cellcolor{gray!25}0.777 &0.015\\
12&\cellcolor{gray!25}0.062&\cellcolor{gray!25}0.424&\cellcolor{gray!25}0.858&\cellcolor{gray!25}0.798&\cellcolor{gray!25}0.711 &0.042\\
13&\cellcolor{gray!25}0.164 &0&\cellcolor{gray!25}0.463&\cellcolor{gray!25}0.438&\cellcolor{gray!25}0.235&\cellcolor{gray!25}0.214\\
14 &0&\cellcolor{gray!25}0.553&\cellcolor{gray!25}0.183&\cellcolor{gray!25}0.701&\cellcolor{gray!25}0.121&\cellcolor{gray!25}0.549\\
15&\cellcolor{gray!25}0.209&\cellcolor{gray!25}0.928 &0.001&\cellcolor{gray!25}0.431&\cellcolor{gray!25}0.45&\cellcolor{gray!25}0.527\\
16 &0.017&\cellcolor{gray!25}0.534&\cellcolor{gray!25}0.259&\cellcolor{gray!25}0.81&\cellcolor{gray!25}0.712&\cellcolor{gray!25}0.662\\
17&\cellcolor{gray!25}0.514&\cellcolor{gray!25}0.276&\cellcolor{gray!25}0.877&\cellcolor{gray!25}0.528&\cellcolor{gray!25}0.545&\cellcolor{gray!25}0.699\\
\hline
KSp & 0 & 0 & 0.504 & 0.340 & 0.787 & 0.126 \\
\end{tabular}
    \label{tab:decres}
\end{table*}

\section{Discussion}\label{sec:disc}

The rules presented in this work provide a guideline to researchers which causal statements are and which ones are not supported by empirical data when analyzing encoding and decoding models. In particular, we argued that only encoding models in stimulus-based paradigms support unambiguous causal statements.
We demonstrated that further causal insights can be derived by combining encoding and decoding models, and illustrated the significance of this theoretical result on experimental data. While we have chosen an EEG study for this illustration, our causal interpretation rules apply to any type of brain state feature. In the following, we discuss limitations and potential extensions of this framework.

\subsection{The iid assumption in neuroimaging}

It is likely that neither of the iid assumptions is met by neuroimaging data.
Consider for example subjects getting tired through the course of an experiment. In this case, the features for later trials may follow another distribution than those for earlier trials, violating the assumption of identical distributions. Also the brain's state may depend on previous activations and hence later trials may be dependent on previous trials, violating the independence assumption. As statistical tests rest on the iid assumption, it is important to consider this limitation when interpreting test results.

\subsection{Finite empirical data}

Our theoretical arguments rest on the assumption that it is possible to distinguish between relevant and irrelevant features in encoding and decoding models, i.\,e.~that we have access to an oracle for univariate independence and multivariate conditional independence tests. In practice, we are faced with several interrelated problems.
Firstly, the identification of irrelevant features rests on the readiness to interpret negative results. We need to interpret the lack of evidence against independence as evidence in favor of it. As such, we need to keep in mind that observing more data and/or using more powerful (conditional) independence tests may falsify previous statistical tests, thereby altering our causal conclusions.
Secondly, we either need to employ non-linear encoding and decoding models to test for (conditional) independences, as in the analysis of our experimental data in section \ref{sec:expres}, or we need to assume that all observed brain state features are jointly Gaussian. This follows from the fact that linear models only test for uncorrelatedness, and in general uncorrelatedness only implies independence if all variables are jointly Gaussian.
Note that other frameworks enable causal discovery in linear models, e.g.~by introducing the additional assumption of additive non-Gaussian noise \cite{shimizu2006linear,hoyer2008causal}.
Lastly, it is difficult to base conditional independence tests on permutation approaches, as these are biased towards conditional dependence \cite{strobl2008conditional}. The development of unbiased conditional independence tests is an area of active research \cite{KunKernel}.

\subsection{Univariate vs multivariate analysis}

We based our causal analysis on commonly employed and intuitive notions of feature relevance (cf.~section \ref{sec:interrel}). In encoding models, a feature is relevant if it varies with the experimental condition. This corresponds to an univariate independence test $X_i \indep C$. In decoding models, a feature is relevant if it cannot be removed without increasing the minimum Bayes error. This corresponds to a multivariate conditional independence test $X_i \indep C | X \setminus X_i$. The interpretation rules presented in this work apply whenever these tests are employed, independently of whether encoding and decoding models or direct statistical tests for (conditional) independence are being used \cite{gretton2008kernel,KunKernel}.

We note that there are instances in which decoding models are used to carry out an encoding analysis. Consider the searchlight technique \cite{kriegeskorte2006information}. Here, it is tested whether a set of $k$ voxels as a whole contains information about the experimental condition. In this case, the decoding model is used for a marginal independence test $C \indep \{X_{i_1},...,X_{i_k}\}$. This approach is oblivious to the causal structure within the set $\{X_{i_1},...,X_{i_k}\}$. As such, the searchlight technique does not provide causal insights beyond those implied by an encoding model. The additional insights offered by decoding models rest on multivariate conditional independence tests.

\subsection{Whole brain analysis}

In case of a high-dimensional feature space, e.\,g.~voxels in fMRI or bandpower features in high-density EEG recordings, training a decoding model on the whole feature set may be infeasible. To harness the additional insights provided by a decoding model, it may be necessary to reduce the feature space dimensionality before training a decoding model, e.\,g.~by clustering for fMRI or ICA for EEG recordings. We note that it is not trivial to reduce the dimensionality of the feature space without discarding causally relevant information.

\subsection{Untestable assumptions}

As it is the case for any type of empirical inference, causal inference also rests on a set of untestable assumptions. In particular, causal inference in the framework of CBNs rests on the CMC and the assumption of faithfulness.
Theoretical results show that the set of unfaithful distributions has measure zero relative to all probability distributions that can be generated by a given DAG \cite{Meek:1995}. As long as nature has no particular reason to favor unfaithful distributions, we are thus unlikely to encounter them in practice. We note that further assumptions might allow stronger causal statements, e.\,g. the assumption of causal sufficiency rules out the existence of hidden common causes.

\subsection{Causal inference in neuroimaging}

We note that the rules presented here have, to a certain extent, already been applied in the context of neuroimaging \cite{Ramsey:2010,Grosse-WentrupNeuroImage2011,waldorp2011effective,mumford2014bayesian}. The primary contribution of our work is to point out their relation to widely used methods for analyzing neuroimaging data. In particular, we show that interpreting encoding and decoding models is a form of causal inference. Given the prevalent use of causal terminology in the interpretation of neuroimaging studies, we believe it is essential to make the inherent assumptions and limitations explicit. We further note that combining encoding and decoding models is only a first step towards a causal analysis of empirical data. More detailed insights can be obtained by additional conditional independence tests, e.\,g.~by training decoding models on subsets of variables and/or permuting subsets of variables. Causal inference algorithms like the PC or FCI algorithm are designed for tackling such questions and hence can yield more detailed causal insights \cite{Pearl2000, Spirtes2000}. In contrast, encoding and decoding models are not especially designed for these purposes but, as shown, might still warrant some causal interpretation.

We close by emphasizing that, if relevant features in encoding and decoding models are interpreted in a causal sense, one inevitably accepts the untestable assumptions and limitations expatiated in this article. The only way to elude this situation, in case one is not willing to make those assumptions, is to resign from causal interpretations.

If not only correlational statements but ultimately neural causes of cognition are of interest, further assumptions and causal inference algorithms, which go beyond encoding and decoding models, should be considered.
Future research may investigate how causal inference methods can be facilitated in neuroimaging, discuss the appropriateness of different assumptions, and explore possible ways to weaken or refine those assumptions in case of neuroimaging data.

\section{Acknowledgment}

The authors want to thank the reviewers for their encouraging, very concise and constructive feedback, which significantly improved this manuscript.

\newpage

\bibliographystyle{plain}
\bibliography{bibfile}

\end{document}